\documentclass{article}
\usepackage[preprint]{neurips_2024}

\usepackage[utf8]{inputenc}
\usepackage[T1]{fontenc}
\usepackage{amsmath,amssymb,amsfonts,amsthm}
\usepackage{booktabs}
\usepackage{array}
\usepackage{tabularx}
\usepackage{enumitem}
\usepackage{algorithm}
\usepackage{algpseudocode}
\usepackage{graphicx}
\usepackage{placeins}
\usepackage{microtype}
\usepackage{url}
\usepackage{hyperref}
\usepackage{natbib}
\usepackage{xcolor}

\hypersetup{
  colorlinks=true,
  linkcolor=blue!45!black,
  citecolor=blue!45!black,
  urlcolor=blue!45!black
}

\newtheorem{theorem}{Theorem}[section]
\newtheorem{proposition}[theorem]{Proposition}
\newtheorem{lemma}[theorem]{Lemma}

\theoremstyle{definition}
\newtheorem{definition}{Definition}
\newtheorem{assumption}{Assumption}
\theoremstyle{remark}

\newcommand{\E}{\mathbb E}
\newcommand{\Pbb}{\mathbb P}
\newcommand{\R}{\mathbb R}
\newcommand{\U}{\mathcal U}
\newcommand{\PiC}{\Pi_{\mathrm{cat}}}
\newcolumntype{Y}{>{\raggedright\arraybackslash}X}

\title{Decision-Calibrated Conformal Uncertainty for Pacing Decisions in Streaming Advertising}

\author{%
  {\fontsize{11}{10}\selectfont Prashant Shekhar\thanks{Corresponding author.} and Caroline Howard} \\
  \textit{\fontsize{10}{22}\selectfont Department of Mathematics} \\
  \textit{\fontsize{10}{22}\selectfont Embry-Riddle Aeronautical University} \\
  \textit{\fontsize{10}{22}\selectfont Daytona Beach, FL, USA}
}

\begin{document}
\maketitle

\begin{abstract}
We develop a decision-calibrated conformal framework for pacing decisions in streaming advertising. Pacing depends on uncertain future inventory, demand pressure, incremental response, and member-experience load. Instead of calibrating a generic forecast residual, the framework measures forecast error by its largest impact on the policies that could actually be deployed. The main theorem shows that the proposed score is the smallest valid uncertainty measure that uniformly protects all deployable pacing policies. Geometrically, it is the support function of the signed policy sensitivity set. Split conformal calibration gives finite-sample coverage for this score. A high-dimensional separation theorem shows that traditional residual calibration can be arbitrarily more conservative by paying for nuisance inventory dimensions, and a robust pacing result combines inventory, response, and experience uncertainty. On public-data-calibrated pacing replays built from Criteo Uplift and KuaiRand datasets, traditional conformal pacing remains unresolved with high residual radii of \(7236.7\) on Criteo and \(4629.4\) on KuaiRand. With the proposed decision calibration approach, the uncertainty radii are reduced to \(18.4\) and \(278.6\) respectively, with separate margins for value, delivery, budget, and member load. On Criteo, the proposed method certifies a less aggressive pacing policy than the point-forecast baseline, and reduces held-out any-violation rate from \(16.7\%\) to \(3.3\%\), with zero budget and member-load violations. On KuaiRand, the choice remains unresolved. Across forecasters, the lowest test MAE doesn't necessarily give the smallest pacing radius. Although a simple temporal-Transformer model attains minimum Criteo MAE, seasonal ridge forecast model attains the smallest uncertainty radius. In a nutshell, the paper establishes that forecasts, response estimates, and member-experience models should be judged by whether they shrink the uncertainty that the pacing decision uses, as this leads to confident decisions that are not overly conservative.
\end{abstract}

\section{Introduction}

Advertising marketplaces are controlled by decisions made before the future is visible. A platform forecasts incoming inventory, paces budgets over time, chooses which campaigns or creatives compete for each opportunity, and enforces product constraints that protect members from excessive ad load or low-quality experiences. These choices are delicate in streaming advertising. Demand may concentrate in particular shows, regions, hours, devices, or audience segments. Campaigns may compete for the same inventory, and a short-run revenue gain can reduce member experience or advertiser incrementality if it overexposes users or exhausts high-quality opportunities too early.

This paper asks how a streaming advertising platform should construct conformal uncertainty sets for pacing decisions when future inventory, incremental response, and member-experience load are all uncertain. Existing work separately studies budgeted ad allocation, pacing equilibria, risk-constrained pacing, neural time-series forecasting, response and policy learning, and robust optimization \citep{mehta2007adwords,devanur2009adwords,balseiro2023robustPacing,gaitonde2022budgetPacing,balseiro2023autobidders,dai2024rcpacing,dudik2011doublyRobust,chernozhukov2018dml,benTal2006robust,bertsimas2011robust}. The gap is that forecasting and pacing methods typically calibrate uncertainty in prediction space, not in the value and constraint directions that determine delivery, budget, value, and member-experience risk.

\begin{figure*}[t]
  \centering
  \includegraphics[width=\textwidth]{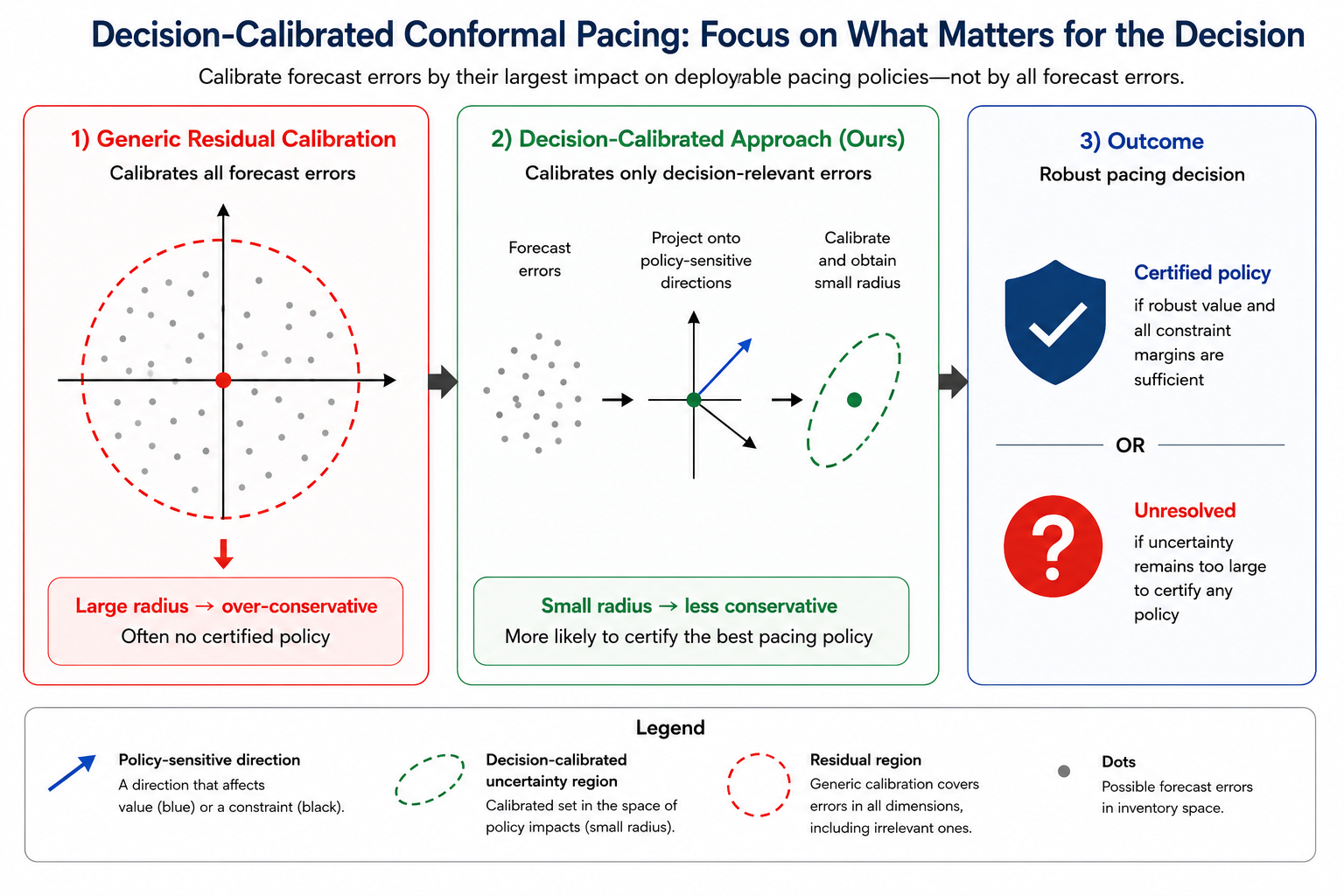}
  \caption{\small Decision-calibrated conformal pacing. Generic residual calibration uses a radius that covers all forecast errors, including irrelevant inventory directions. The decision-calibrated score projects forecast errors onto value and constraint directions used by the pacing catalog, calibrates the resulting policy-impact radius, and returns either a certified policy or an unresolved decision.}
  \label{fig:intro-overview}
\end{figure*}

In one sentence, the paper calibrates forecast uncertainty after projecting forecast errors through the value and constraint directions that matter for the pacing decision. Figure~\ref{fig:intro-overview} shows the central comparison. Generic residual calibration covers all forecast errors and can create a large radius. The method developed here first projects errors onto policy-sensitive directions, then calibrates the resulting policy-impact score. The selector certifies a pacing policy only when the resulting robust value and constraint margins are large enough. Otherwise it returns unresolved.

The key quantity we propose is a \emph{dual-weighted} forecast score. Underforecasting a scarce audience segment needed by a high-value campaign near its delivery constraint can be much more damaging than the same error in slack inventory. The objective gives a value direction, and the pacing dual variables give shadow prices for delivery, budget, and member-experience constraints. Together these directions identify which future inventory errors matter. The score measures the largest policy impact of a forecast error across the deployable catalog. Conformal calibration then controls this decision-relevant error rather than marginal mean absolute error or mean squared error \citep{romano2019cqr,gibbs2021adaptive,xu2020enbpi,angelopoulos2021learnThenTest}. Thus, the optimizer receives uncertainty sets shaped by ad-delivery economics.

The formulation is designed to return the best certified pacing policy. Forecasting enters because future supply and demand are unknown. Response modeling enters because pacing value depends on incremental advertiser response, not raw clicks. Member-experience modeling enters because the selected policy must respect exposure and load constraints. Robust optimization turns these uncertainties into either a certificate or an unresolved shortlist. The finite policy catalog reflects deployment practice, where platforms ship pacing rules, bidding multipliers, and budget controls from auditable catalogs rather than arbitrary measurable policies.

The paper makes three primary contributions. \textbf{First}, it proposes a decision-calibrated conformal pacing method that projects forecast errors through the value and constraint directions used by a deployable policy catalog, calibrates the resulting uncertainty radii, and returns either a certified pacing policy or an unresolved shortlist. \textbf{Second}, it proves that the resulting dual-weighted score is the minimal valid scalar certificate for the catalog and develops finite-sample theory for estimating the value and constraint radii, including a sample-complexity result for stable policy certification. \textbf{Third}, it combines calibrated forecast uncertainty with response and member-experience uncertainty in a robust pacing selector and evaluates the full pipeline on public-data-calibrated Criteo and KuaiRand pacing replays. The experiments report selected policies, realized value, held-out constraint violations, and ablations for geometry, coverage, catalog granularity, response uncertainty, and calibration sample size. Code for reproducing the experiments is available at \url{https://github.com/p-shekhar/pacing-decisions-advertising}.

\section{Related Work}

\noindent \textbf{Budgeted ad allocation and pacing}:
The AdWords problem and generalized online matching formalize online ad allocation under budgeted bidders \citep{mehta2007adwords,devanur2009adwords}. Budget pacing, first-price pacing equilibria, and robust pacing study repeated auctions and budget constraints when demand and bids are uncertain \citep{balseiro2023robustPacing,gaitonde2022budgetPacing,balseiro2023abTestingPacing,balseiro2024interferencePacing}. Autobidding extends this view to budget and return-on-investment constraints \citep{balseiro2023autobidders}. Reinforcement-learning approaches to real-time bidding learn bidding policies in display and sponsored-search systems \citep{zhang2017rtbRL,jin2018multiAgentRTB,cai2018deepRLSponsoredSearch}. Building on that, the current paper asks how to calibrate conformal uncertainty after future inventory, response, and member-experience uncertainty have been translated into value and constraint sensitivity.

\noindent \textbf{Forecasting, predict-then-optimize, and calibrated uncertainty}:
Probabilistic and neural forecasting models such as DeepAR, N-BEATS, Temporal Fusion Transformers, Informer, Autoformer, PatchTST, and foundation-style time-series models provide flexible tools for future demand and inventory \citep{salinas2017deepar,oreshkin2020nbeats,lim2021temporal,zhou2021informer,wu2021autoformer,nie2023patchtst,ansari2024chronos}. Predict-then-optimize and decision-focused learning evaluate predictive models by downstream optimization quality rather than prediction loss alone \citep{elmachtoub2022smart}. Conformalized quantile regression, adaptive conformal inference, ensemble batch prediction intervals, conformal robust optimization, robust contextual linear programming, and utility-directed conformal prediction provide finite-sample calibration tools for predictive decisions \citep{romano2019cqr,gibbs2021adaptive,xu2020enbpi,angelopoulos2021learnThenTest,johnstone2021conformalRobust,sun2023predictCalibrate,cortesGomez2025utilityDirected}. We treat the forecasting model as an upstream input. The contribution is to move calibration from prediction space into the directions that matter for pacing. The empirical forecasting comparison tests whether prediction accuracy and decision radius can disagree.

\noindent \textbf{Incremental response as an input to pacing}:
Logged ad and recommendation systems produce biased samples of actions and outcomes. Doubly robust policy learning, counterfactual risk minimization, double machine learning, causal forests, and representation learning for treatment effects provide tools for estimating incremental response and policy value from logged data \citep{dudik2011doublyRobust,swaminathan2015crm,chernozhukov2018dml,athey2019grf,shalit2017estimating}. Public datasets such as Criteo Uplift, Open Bandit, and KuaiRand make related empirical tests possible \citep{diemert2018criteo,saito2020openBandit,gao2022kuairand}. We use these estimates as inputs to the pacing certificate. The methodological contribution is the uncertainty layer that decides how forecast, response, and experience uncertainty should jointly affect a pacing decision.

\noindent \textbf{Robust optimization and constrained online learning}:
Robust optimization studies decisions that remain feasible under uncertainty sets \citep{benTal2006robust,bertsimas2011robust}. Bandits with knapsacks and online convex optimization with long-term constraints study online decisions under resource constraints \citep{badanidiyuru2013bwk,jenatton2016ocoConstraints}. We use this perspective to derive decision guarantees for ad pacing policies selected from calibrated forecast and response uncertainty sets.

\noindent \textbf{Closest distinction}:
Closest to our setting are conformal robust optimization, robust contextual linear programming, and utility-directed conformal prediction \citep{johnstone2021conformalRobust,sun2023predictCalibrate,cortesGomez2025utilityDirected}. Those papers establish decision-aware calibration as a general methodology. Our contribution is the pacing-specific geometry. Objective sensitivity together with delivery, budget, and member-experience dual prices induces the minimal valid certificate for a streaming ad-delivery catalog. The same geometry exposes a failure mode of generic residual calibration, which can reserve slack for nuisance inventory dimensions that are invisible to every deployable pacing policy.

\section{Problem Setup}

Time is indexed by \(t=1,\ldots,T\). At each time, a streaming platform receives ad opportunities with observable context, such as region, device, title, hour, user segment, and campaign eligibility. There are \(K\) campaigns. Campaign \(k\) has delivery target \(D_k\), budget \(B_k\), and value-per-incremental-outcome \(v_k\). A platform action may include the displayed campaign, creative, bid multiplier, reserve adjustment, or no-ad decision.

Let \(\Pi\) be the class of feasible pacing policies, and let \(\pi_t\) be the action selected by policy \(\pi\) at time \(t\). For campaign \(k\), let \(\tau_{k,t}(\pi_t)\) denote incremental advertiser response, \(d_{k,t}(\pi_t)\) denote delivered units, and \(c_{k,t}(\pi_t)\) denote budget spend. The response \(\tau_{k,t}(\pi_t)\) is distinct from raw click-through or conversion probability because it measures incremental value relative to not showing the ad or showing an alternative. Let \(Q_t(\pi_t)\) denote member-experience load or quality cost, let \(Q_{\max}\) be the allowed total member-experience load, and let \(\lambda_{\mathrm{load}}\ge 0\) be the objective penalty on member load. With these notations defined (also summarized in Table \ref{tab:setup-symbol-guide}), in the following formulation, expectations are placed over future opportunity arrivals and outcomes. The notation \(\E_\pi\) emphasizes the distribution induced by policy \(\pi\).

\begin{table}[t]
\centering
\small
\caption{\small Problem-setup notation used in the objective and constraints.}
\label{tab:setup-symbol-guide}
\begin{tabularx}{0.92\linewidth}{@{}>{\raggedright\arraybackslash}p{0.20\linewidth}X@{}}
\toprule
Symbol & Meaning \\
\midrule
\(\pi\), \(\pi_t\) & Pacing policy and the action it selects at time \(t\). \\
\(D_k\) & Delivery target for campaign \(k\). \\
\(B_k\) & Budget for campaign \(k\). \\
\(v_k\) & Value per incremental outcome for campaign \(k\). \\
\(\tau_{k,t}(\pi_t)\) & Incremental advertiser response under the selected action. \\
\(d_{k,t}(\pi_t)\) & Delivered units for campaign \(k\). \\
\(c_{k,t}(\pi_t)\) & Budget spend for campaign \(k\). \\
\(Q_t(\pi_t)\) & Member-experience load or quality cost. \\
\(Q_{\max}\) & Allowed total member-experience load. \\
\bottomrule
\end{tabularx}
\end{table}

The clean objective is
\begin{equation}
\label{eq:clean-objective}
\max_{\pi\in\Pi}
\E\!\left[
\sum_{t=1}^T
\sum_{k=1}^K
v_k\,\tau_{k,t}(\pi_t)
- \lambda_{\mathrm{load}} Q_t(\pi_t)
\right],
\end{equation}
subject to campaign delivery, budget, and member-experience constraints.
\begin{align}
\label{eq:delivery-constraints}
\E_{\pi}\!\left[\sum_{t=1}^T d_{k,t}(\pi_t)\right] &\ge D_k,
&& k=1,\ldots,K,\\
\label{eq:budget-constraints}
\E_{\pi}\!\left[\sum_{t=1}^T c_{k,t}(\pi_t)\right] &\le B_k,
&& k=1,\ldots,K,\\
\label{eq:member-constraint}
\E_{\pi}\!\left[\sum_{t=1}^T Q_t(\pi_t)\right] &\le Q_{\max}.
\end{align}
The difficulty is that future opportunity volume, \(\tau_{k,t}(\cdot)\), and parts of \(Q_t(\cdot)\) are not known before launch. The paper therefore replaces point optimization with robust optimization over calibrated uncertainty sets.

\section{Decision-Calibrated Conformal Pacing}

\subsection{Forecasting as an input to pacing uncertainty}

Let \(z\) denote the vector of future quantities over the planning horizon. In the implementation, each segment-block vector has coordinates for inventory volume, eligible demand pressure, and quality-adjusted opportunity mass. A forecasting model fitted on historical logs produces the point forecast \(\widehat z\).

A raw forecast residual treats every coordinate of \(z\) as equally important. A pacing decision does not. If the forecast is wrong for an audience segment that no active campaign can use, the error may have little effect on delivery, budget, value, or member load. If the same-sized error occurs in scarce inventory needed by a high-value campaign near a delivery constraint, it can change which policy should be launched. The question is therefore not only how large the forecast error is, but whether the error falls in directions that affect the pacing decision.

Different inventory dimensions matter for different reasons. Some coordinates affect expected value because they create high-quality advertising opportunities. Others affect feasibility because they determine whether campaigns can meet delivery goals, stay within budget, or respect member-experience limits. A useful uncertainty score should reflect these differences. Reducing prediction error in an irrelevant coordinate should not force the platform to reserve more pacing slack, while reducing error in a scarce, constraint-relevant coordinate should matter.

For each deployable policy \(\pi\), we summarize this dependence by a sensitivity vector \(w_\pi\).Conceptually, \(w_\pi\) measures how the downstream pacing quantities for policy \(\pi\) respond to changes in the forecasted planning vector, including changes in value and constraint margins. The forecast error enters only after this sensitivity is paired with \(e=z-\widehat z\). Large entries of \(w_\pi\) mark forecast coordinates whose errors can change the decision. Small or zero entries mark coordinates that are mostly irrelevant for that policy. Let \(\PiC\subseteq\Pi\) denote the finite deployable policy catalog, and let \(J\) be the number of constraint functions. The Lagrangian below is only a bookkeeping device for computing these sensitivity directions. It combines value sensitivity with the delivery, budget, and member-experience dual prices from the nominal relaxation, producing relevant directions for each policy in the catalog. The forecasting model itself remains model-agnostic and it may be simple or neural, but it is judged by whether its errors are small in these policy-relevant directions.

\begin{definition}[Decision-calibrated forecast score]
\label{def:dc-score}
Let \(\widehat u=(\widehat z,\widehat \tau,\widehat m)\) be the nominal future world, collecting inventory, incremental-response, and member-experience forecasts, and let \(\widehat\lambda\ge 0\) be a vector of dual prices for the delivery, budget, and member-experience constraints in the nominal pacing relaxation. For policy \(\pi\), define the Lagrangian
\[
\mathcal L(\pi;u,\widehat\lambda)
=
V(\pi;u)-\sum_{j=1}^J \widehat\lambda_j g_j(\pi;u),
\]
where \(V\) is platform value and \(g_j\le 0\) are constraint violations. The inventory sensitivity of \(\pi\) is
\[
w_\pi(\widehat u,\widehat\lambda)
=
\nabla_z \mathcal L(\pi;\widehat u,\widehat\lambda).
\]
For calibration block \(b\), with realized future \(z_b\) and forecast \(\widehat z_b\), the dual-weighted forecast score is
\[
S_b^{\mathrm{dc}}
=
\max_{\pi\in\PiC}
\left|
\left\langle
w_\pi(\widehat u_b,\widehat\lambda_b),
z_b-\widehat z_b
\right\rangle
\right|.
\]
\end{definition}

For a forecast error \(e=z-\widehat z\), write the corresponding catalog support function as
\begin{equation}
\label{eq:support-score}
\Phi_{\PiC}(e;\widehat u,\widehat\lambda)
=
\max_{\pi\in\PiC}
\left|
\left\langle
w_\pi(\widehat u,\widehat\lambda),e
\right\rangle
\right|.
\end{equation}
Then the calibration score is \(S_b^{\mathrm{dc}}=\Phi_{\PiC}(z_b-\widehat z_b;\widehat u_b,\widehat\lambda_b)\). It is large when the forecast error falls in inventory directions with high downstream value or high constraint pressure. A forecaster improves the pacing decision when it reduces this score, even if another model has lower generic prediction error. Given calibration scores \(S_1^{\mathrm{dc}},\ldots,S_n^{\mathrm{dc}}\) and miscoverage level \(\alpha\), define the split-conformal quantile
\begin{equation}
\label{eq:dc-quantile}
\widehat q^{\mathrm{dc}}_{1-\alpha}
=
\text{the }
\left\lceil (n+1)(1-\alpha)\right\rceil
\text{th order statistic of }
\{S_b^{\mathrm{dc}}\}_{b=1}^n .
\end{equation}

\begin{assumption}[Exchangeable calibration blocks]
\label{ass:exchange}
The calibration score blocks and the future test block are exchangeable after conditioning on the training procedure used to fit the forecasting model and compute the nominal pacing dual prices.
\end{assumption}

\begin{assumption}[Affine downstream inventory contribution]
\label{ass:affine}
For every \(\pi\in\PiC\), the inventory-dependent part of the evaluated Lagrangian is affine in \(z\) over the planning block, with coefficient \(w_\pi(\widehat u,\widehat\lambda)\). When the downstream functions are differentiable rather than exactly affine, the proof applies to the first-order term and adds the usual second-order Taylor remainder.
\end{assumption}

The next theorem gives the score its decision meaning. For a forecast error \(e\), it asks how much any deployable policy's downstream value-and-constraint tradeoff can change. The answer is the largest inner product between \(e\) and the catalog sensitivities. This is the quantity to control if one radius must protect every candidate policy.

The same quantity has a geometric form. The catalog induces sensitivity vectors \(w_\pi\) and their negatives, since both over- and under-forecasting can hurt. Their convex hull contains every signed catalog direction. The largest policy effect is the support function of this hull. Any valid scalar certificate for the catalog must contain the same geometry.

\begin{theorem}[Decision-calibrated uncertainty characterization]
\label{thm:sharp-geometry}
\mbox{}\\[-0.5ex]
The result has four parts, an operational identity, a geometric representation, a minimality characterization, and a conformal coverage statement.

\textbf{Part I. Operational characterization}:\par
Suppose Assumption~\ref{ass:affine} holds. Fix a nominal world \((\widehat u,\widehat\lambda)\). For any forecast error \(e=z-\widehat z\), the largest downstream Lagrangian change over the deployable pacing catalog is
\[
\sup_{\pi\in\PiC}
\left|
\mathcal L(\pi;(\widehat z+e,\widehat\tau,\widehat m),\widehat\lambda)
-
\mathcal L(\pi;(\widehat z,\widehat\tau,\widehat m),\widehat\lambda)
\right|
=
\Phi_{\PiC}(e;\widehat u,\widehat\lambda).
\]

\textbf{Part II. Geometric representation}:\par
Now define the signed policy-sensitivity set
\[
\mathcal W_{\PiC}(\widehat u,\widehat\lambda)
=
\operatorname{conv}
\left(
\{w_\pi(\widehat u,\widehat\lambda):\pi\in\PiC\}
\cup
\{-w_\pi(\widehat u,\widehat\lambda):\pi\in\PiC\}
\right).
\]
The same catalog effect has the support-function representation
\[
\Phi_{\PiC}(e;\widehat u,\widehat\lambda)
=
\sigma_{\mathcal W_{\PiC}}(e)
:=
\sup_{v\in\mathcal W_{\PiC}(\widehat u,\widehat\lambda)}
\langle v,e\rangle .
\]

\textbf{Part III. Minimality characterization}:\par
This part shows that any valid scalar certificate must cover every signed sensitivity direction generated by the catalog.
Geometrically, any valid uncertainty measure must contain all directions that can affect a deployable policy.
Consider any finite-valued lower-semicontinuous sublinear scalar certificate \(B\), meaning that \(B\) is nonnegative, positively homogeneous, and subadditive. If \(B\) uniformly certifies all catalog Lagrangian errors,
\[
\left|
\mathcal L(\pi;(\widehat z+e,\widehat\tau,\widehat m),\widehat\lambda)
-
\mathcal L(\pi;(\widehat z,\widehat\tau,\widehat m),\widehat\lambda)
\right|
\le B(e)
\quad
\text{for all }\pi\in\PiC\text{ and all }e,
\]
then there exists a closed convex set
\[
\mathcal K_B
=
\{v:\langle v,e\rangle\le B(e)\ \text{for all }e\}
\]
such that \(B=\sigma_{\mathcal K_B}\) and
\[
\mathcal W_{\PiC}(\widehat u,\widehat\lambda)\subseteq \mathcal K_B .
\]
Consequently \(B(e)\ge \Phi_{\PiC}(e;\widehat u,\widehat\lambda)\) for all \(e\). Equality for all \(e\) holds if and only if \(\mathcal K_B=\mathcal W_{\PiC}\). Thus the score is the unique minimal coherent scalar certificate for the catalog.

\textbf{Part IV. Conformal validity}:\par
If Assumption~\ref{ass:exchange} also holds and \(\widehat q^{\mathrm{dc}}_{1-\alpha}\) is the conformal quantile in Eq.~\eqref{eq:dc-quantile}, then for the next planning block,
\[
\Pbb\!\left(
\max_{\pi\in\PiC}
\left|
\left\langle
w_\pi(\widehat u,\widehat\lambda),
z-\widehat z
\right\rangle
\right|
\le
\widehat q^{\mathrm{dc}}_{1-\alpha}
\right)
\ge
1-\alpha.
\]
\end{theorem}

Theorem~\ref{thm:sharp-geometry} shows that the relevant uncertainty is not the full forecast residual, but the part seen by the current pacing catalog. The minimality result rules out smaller valid scalar certificates under catalog-level robustness. A smaller radius requires additional structure beyond the catalog sensitivities. The conformal statement makes the score usable with any upstream forecaster, while keeping the radius in downstream pacing units. Appendix Figure~\ref{fig:appendix-support-characterization} gives the corresponding diagnostic. Strict supersets of the signed sensitivity hull are valid but conservative, while omitting an active direction or scaling below the support function breaks catalog validity.

The selector applies the same calibration recipe more than once. The Lagrangian score gives one radius for the combined value-and-constraint tradeoff, which is useful for the geometry theorem. Deployment also needs component-level statements. The platform needs a lower confidence bound on value and upper confidence bounds on delivery shortfall, budget overspend, and member-experience load. A single radius would hide which part of the decision is safe and which part is binding.

The implementation therefore calibrates value and each constraint separately. Each component has a sensitivity direction, a calibration score, and a split-conformal radius. The value sensitivity measures how platform value responds locally to changes in the forecasted planning vector. The constraint sensitivities measure how each feasibility margin responds locally to the same forecast coordinates. Forecast error is then evaluated by pairing these sensitivities with \(e=z-\widehat z\). The resulting radii enter the robust selector as value discounts and constraint tightenings.

The following scores are direct analogues of \(S_b^{\mathrm{dc}}\), but applied to the individual decision components rather than to the combined Lagrangian direction. Let
\[
a_{\pi,V}=\nabla_z V(\pi;\widehat u),
\qquad
a_{\pi,g_j}=\nabla_z g_j(\pi;\widehat u),\qquad j=1,\ldots,J.
\]
For calibration block \(b\), define
\[
S_b^V=\max_{\pi\in\PiC}|\langle a_{\pi,V},z_b-\widehat z_b\rangle|,
\qquad
S_b^{g_j}=\max_{\pi\in\PiC}|\langle a_{\pi,g_j},z_b-\widehat z_b\rangle|.
\]
Let \(\alpha_{\mathrm{comp}}\) denote the component miscoverage level used for value and constraint certification. The corresponding split-conformal quantiles are denoted
\[
\widehat q^V_{1-\alpha_{\mathrm{comp}}},
\qquad
\widehat q^{g_j}_{1-\alpha_{\mathrm{comp}}},\quad j=1,\ldots,J.
\]
In the experiments, \(J=3\), corresponding to delivery shortfall, budget overspend, and member-experience load, and Bonferroni calibration sets \(\alpha_{\mathrm{comp}}=\alpha/(J+1)=0.025\).

\begin{table}[t]
\centering
\small
\caption{\small Guide to the main calibration quantities. The first three rows define the combined Lagrangian score used in Theorem~\ref{thm:sharp-geometry}. The remaining rows define the component radii used by the deployable selector in Eq.~\eqref{eq:robust-policy}.}
\label{tab:symbol-guide}
\begin{tabularx}{\linewidth}{@{}>{\raggedright\arraybackslash}p{0.23\linewidth}>{\raggedright\arraybackslash}p{0.33\linewidth}X@{}}
\toprule
Symbol & Plain-language meaning & Role \\
\midrule
\(w_\pi\) & Sensitivity direction for policy \(\pi\) & Prices forecast errors by how they affect value and constraint margins. \\
\(S_b^{\mathrm{dc}}\) & Combined decision score on calibration block \(b\) & Measures the largest Lagrangian error across the catalog. \\
\(\widehat q^{\mathrm{dc}}_{1-\alpha}\) & Conformal radius for the combined score & Certifies the catalog-level Lagrangian error in Theorem~\ref{thm:sharp-geometry}. \\
\(S_b^V\) & Value-only calibration score & Measures how forecast error changes platform value. \\
\(S_b^{g_j}\) & Constraint-\(j\) calibration score & Measures how forecast error changes delivery, budget, or member-load margin \(g_j\). \\
\(\widehat q^V_{1-\alpha_{\mathrm{comp}}}\) & Value forecast radius & Subtracted from nominal value in Eq.~\eqref{eq:robust-policy}. \\
\(\widehat q^{g_j}_{1-\alpha_{\mathrm{comp}}}\) & Constraint-\(j\) forecast radius & Added to constraint \(g_j\) in Eq.~\eqref{eq:robust-policy}. \\
\(\rho_V(\pi)\) & Response uncertainty radius for policy value & Further discounts value when incremental response is uncertain. \\
\(\rho_{g,j}(\pi)\) & Response or experience radius for constraint \(j\) & Further tightens constraints when response or member-experience estimates are uncertain. \\
\bottomrule
\end{tabularx}
\end{table}

\subsection{Response and experience uncertainty in the pacing certificate}

For each action \(a\), let \(\tau_t(a)\) be incremental advertiser response and \(m_t(a)\) be member-experience cost. The platform estimates these functions from randomized experiments, logged bandit data, or quasi-experimental variation. We represent uncertainty through intervals
\[
\tau_t(a)\in [\widehat\tau_t(a)-\rho_{\tau,t}(a),\widehat\tau_t(a)+\rho_{\tau,t}(a)],
\qquad
m_t(a)\in [\widehat m_t(a)-\rho_{m,t}(a),\widehat m_t(a)+\rho_{m,t}(a)].
\]
The uncertainty radius may come from doubly robust scores, cross-fitting, causal forests, conformal risk control, or domain-specific sensitivity analysis. The framework does not require a new response estimator. It requires intervals that can be used by the pacing decision. In the implementation, response uncertainty contributes to the policy-level value radius \(\rho_V(\pi)\), member-experience uncertainty contributes to the member-load constraint radius, and delivery and budget constraints receive no additional response or experience radius beyond their calibrated forecast component radii. More generally, these intervals can define policy-level value and constraint radii \(\rho_V(\pi)\) and \(\rho_{g,j}(\pi)\).

The robust selector discounts value by the forecast-calibration margin \(\widehat q^V\) and the response margin \(\rho_V(\pi)\). It tightens each constraint by its forecast-calibration margin \(\widehat q^{g_j}\) and any component-specific response or experience margin \(\rho_{g,j}(\pi)\). A policy is certified only if it remains feasible after these adjustments. The selected policy maximizes the resulting value lower bound subject to the corresponding constraint upper bounds.
\begin{equation}
\label{eq:robust-policy}
\widehat\pi
\in
\arg\max_{\pi\in\PiC}
\left\{
\widehat V(\pi)-\widehat q^V_{1-\alpha_{\mathrm{comp}}}-\rho_V(\pi)
\right\}
\quad
\text{subject to}
\quad
\widehat g_j(\pi)+\widehat q^{g_j}_{1-\alpha_{\mathrm{comp}}}+\rho_{g,j}(\pi)\le 0,\ j=1,\ldots,J.
\end{equation}
Here \(\widehat V\) and \(\widehat g_j\) are evaluated at the point forecast and estimated response functions. The component radii \(\widehat q^V\) and \(\widehat q^{g_j}\) use the same support-function construction as \(\widehat q^{\mathrm{dc}}\), applied to the value and constraint directions certified by the selector. This is the implementation used in the experiments.

The implementation (Algorithm \ref{alg:selector}) also reports an uncertainty-aware shortlist. Let \(\widehat R^\star\) be the best robust value among certified policies, or the best robust value over the full catalog if no policy is certified. Given a planning tolerance \(\eta_{\mathrm{plan}}\), the shortlist contains policies whose robust value is at least \(\widehat R^\star-\eta_{\mathrm{plan}}|\widehat R^\star|\). All experiments use \(\eta_{\mathrm{plan}}=0.05\).

\begin{algorithm}[t]
\caption{Decision-Calibrated Conformal Pacing Selector}
\label{alg:selector}
\begin{algorithmic}[1]
\Require Historical logs, policy catalog \(\PiC\), campaign constraints, miscoverage levels \(\alpha,\alpha_\tau,\alpha_m\), planning tolerance \(\eta_{\mathrm{plan}}\).
\Ensure Selected pacing policy \(\widehat\pi\), robust value, constraint certificate, and uncertainty shortlist.
\State Split the streaming blocks into training, calibration, and held-out planning blocks.
\State Fit a forecasting model on the training blocks to predict future inventory, demand pressure, and opportunity quality \(z\).
\State On each calibration block \(b\), compute forecast error \(e_b=z_b-\widehat z_b\).
\State Solve finite-catalog linear relaxation of Eqs.~\eqref{eq:clean-objective}--\eqref{eq:member-constraint} at calibration forecast \(\widehat z_b\), using mixture weights over \(\Pi_{\mathrm{cat}}\), and record delivery, budget, and member-experience dual prices \(\widehat\lambda_b\) returned by the linear program.

\State Compute the Lagrangian support score \(S_b^{\mathrm{dc}}\) from Definition~\ref{def:dc-score}, equivalently Eq.~\eqref{eq:support-score}.
\State Compute the Lagrangian decision radius \(\widehat q^{\mathrm{dc}}_{1-\alpha}\) using Eq.~\eqref{eq:dc-quantile}.
\State For value certification, compute \(S_b^V=\max_{\pi\in\PiC}|\langle a_{\pi,V},e_b\rangle|\) and its conformal radius \(\widehat q^V_{1-\alpha_{\mathrm{comp}}}\) using the same order statistic as Eq.~\eqref{eq:dc-quantile}.
\State For each constraint \(j=1,\ldots,J\), compute \(S_b^{g_j}=\max_{\pi\in\PiC}|\langle a_{\pi,g_j},e_b\rangle|\) and its conformal radius \(\widehat q^{g_j}_{1-\alpha_{\mathrm{comp}}}\) using the same order statistic as Eq.~\eqref{eq:dc-quantile}.
\State Estimate response and member-experience uncertainty. In the implemented experiments, convert response uncertainty to \(\rho_V(\pi)\), convert member-experience uncertainty to the member-load constraint radius, and set additional response/experience radii for delivery and budget constraints to zero.
\State Form the planning forecast \(\widehat z_{\mathrm{plan}}\) from the mean predicted held-out horizon generated by the rolling replay.
\For{each policy \(\pi\in\PiC\)}
  \State Evaluate nominal \(\widehat V(\pi)\) and \(\widehat g_j(\pi)\) at \(\widehat z_{\mathrm{plan}}\).
  \State Compute robust value lower bound \(\widehat R(\pi)=\widehat V(\pi)-\widehat q^V_{1-\alpha_{\mathrm{comp}}}-\rho_V(\pi)\), as in Eq.~\eqref{eq:robust-policy}.
  \State Compute robust constraint upper bounds \(\widehat G_j(\pi)=\widehat g_j(\pi)+\widehat q^{g_j}_{1-\alpha_{\mathrm{comp}}}+\rho_{g,j}(\pi)\), as in Eq.~\eqref{eq:robust-policy}.
  \State Mark \(\pi\) certified if \(\widehat G_j(\pi)\le 0\) for all \(j=1,\ldots,J\).
\EndFor
\State Select \(\widehat\pi\) as the certified policy with largest \(\widehat R(\pi)\), matching Eq.~\eqref{eq:robust-policy}.
\State Let \(\widehat R^\star\) be the best robust value among certified policies, or the best robust value over \(\PiC\) if no policy is certified.
\State Return all policies with \(\widehat R(\pi)\ge \widehat R^\star-\eta_{\mathrm{plan}}|\widehat R^\star|\). If no policy is certified, return this as an unresolved shortlist.
\end{algorithmic}
\end{algorithm}

\section{Theoretical Results}

The remaining main results explain how the calibrated radii behave in deployment. They address how many calibration blocks are needed, why nuisance inventory dimensions can make generic residual radii too large, and how forecast, response, and experience uncertainty combine in the final certificate.

The first result turns conformal coverage into a calibration-sample-size rule for stable policy certification.

\begin{proposition}[Calibration sample size for stable pacing certification]
\label{prop:calibration-sample-size}
Let \(K_{\mathrm{comp}}=J+1\) be the number of calibrated decision components used by Eq.~\eqref{eq:robust-policy}, consisting of one value component and \(J\) constraint components. For component \(k\in\{V,g_1,\ldots,g_J\}\), let \(S_k\) be its score, let \(F_k\) be the population distribution of \(S_k\), and let
$
q_k=F_k^{-1}(1-\alpha_k)
$
be the population calibration radius. Suppose that, for some \(f_{\min}>0\) and \(\varepsilon>0\), every \(F_k\) has density at least \(f_{\min}\) on
$
[q_k-\varepsilon,\ q_k+\varepsilon].
$
Let \(\widehat q_k\) be the split-conformal empirical quantile computed from \(n_{\mathrm{cal}}\) exchangeable calibration blocks. Then, there is a universal constant \(C>0\) such that if
\[
n_{\mathrm{cal}}
\ge
\frac{C}{f_{\min}^2\varepsilon^2}
\log\frac{2K_{\mathrm{comp}}}{\delta},
\]
then, with probability at least \(1-\delta\),
\[
\max_{k\in\{V,g_1,\ldots,g_J\}}
|\widehat q_k-q_k|
\le
\varepsilon .
\]
Consequently, every robust value lower bound and every robust constraint upper bound used by Eq.~\eqref{eq:robust-policy} is within \(\varepsilon\) of its population-calibrated counterpart, uniformly over the policy catalog. If the population-calibrated selector has a unique feasible optimizer whose active feasibility and value margins exceed \(2\varepsilon\), then the empirical selector returns the same policy. When these margins are smaller, calibration uncertainty is large enough to affect the decision, and the appropriate output is an uncertainty-aware shortlist rather than a forced recommendation.
\end{proposition}

Proposition~\ref{prop:calibration-sample-size} gives the sample-complexity layer. A platform does not need enough historical blocks to estimate a band in every nuisance inventory coordinate. It needs enough calibration blocks to estimate the high quantile of the decision score. The local density condition is the usual quantile stability requirement. If many scores pile up near the target quantile, the radius is hard to estimate. If the distribution crosses the quantile cleanly, fewer blocks are needed.

The next result shows why this distinction matters when many forecast coordinates are irrelevant to the current pacing catalog.

\begin{theorem}[High-dimensional separation from generic residual calibration]
\label{thm:separation}
Let the inventory-forecast space decompose orthogonally as \(\R^d=\mathcal S\oplus \mathcal N\), where all catalog sensitivity vectors satisfy \(w_\pi\in\mathcal S\) and \(\mathcal N\) is a nuisance subspace that no pacing policy prices. Then for any forecast error \(e=e_{\mathcal S}+e_{\mathcal N}\),
\[
\Phi_{\PiC}(e;\widehat u,\widehat\lambda)
=
\Phi_{\PiC}(e_{\mathcal S};\widehat u,\widehat\lambda),
\]
whereas a generic Euclidean residual score depends on both components.
\[
\|e\|_2^2=\|e_{\mathcal S}\|_2^2+\|e_{\mathcal N}\|_2^2.
\]
Consequently, for every \(A>1\), there exists an exchangeable calibration/test distribution with valid conformal radii \(q^{\mathrm{dc}}_{1-\alpha}\) and \(q^{\mathrm{gen}}_{1-\alpha}\) such that
\[
q^{\mathrm{gen}}_{1-\alpha}
\ge
A\,q^{\mathrm{dc}}_{1-\alpha},
\]
while \(q^{\mathrm{dc}}_{1-\alpha}\) remains the sharp radius for all catalog Lagrangian forecast errors. More concretely, in dimension \(d=m+1\), a one-policy catalog with sensitivity \(w=e_1\), where \(e_1\) is the first coordinate vector, and exchangeable forecast errors with \(\sigma>0\)
\[
e=(\xi_0,\sigma \xi_1,\ldots,\sigma \xi_m),
\qquad
\xi_0,\xi_1,\ldots,\xi_m\in\{-1,1\},
\]
with independent Rademacher signs,
has
\[
q^{\mathrm{dc}}_{1-\alpha}=1,
\qquad
q^{\mathrm{gen}}_{1-\alpha}=\sqrt{1+m\sigma^2}.
\]
Thus the conservatism of norm-based residual calibration grows as \(\sqrt{m}\) when nuisance inventory error has constant scale.
\end{theorem}

Theorem~\ref{thm:separation} gives a failure mode for accuracy-first uncertainty quantification. A forecaster may err in inventory coordinates that do not affect the current campaign mix, while a generic residual interval still forces the optimizer to reserve slack for those errors. The dual-weighted score avoids this by projecting forecast error through the catalog sensitivities and discarding nuisance directions with zero price for the current decision. The calibration is therefore not just a model-comparison exercise. It changes which uncertainty can constrain the business decision.

The final main theorem combines the calibrated inventory radii with response and member-experience uncertainty in the selector.

\begin{assumption}[Response and experience coverage]
\label{ass:responsecoverage}
Let \(u^\star\) be the realized future world, and let \(\alpha_\tau\) and \(\alpha_m\) be the response and member-experience miscoverage levels. Let \(\alpha_{\mathrm{comp}}\) be the component-level conformal miscoverage used to calibrate the value and constraint forecast radii. With probability at least \(1-\alpha-\alpha_\tau-\alpha_m\), the calibrated forecast radii and the estimated response and member-experience radii satisfy, for all \(\pi\in\PiC\) and all constraints \(j\),
\[
|V(\pi;u^\star)-\widehat V(\pi)|\le
\widehat q^V_{1-\alpha_{\mathrm{comp}}}+\rho_V(\pi),
\qquad
g_j(\pi;u^\star)\le
\widehat g_j(\pi)+\widehat q^{g_j}_{1-\alpha_{\mathrm{comp}}}+\rho_{g,j}(\pi).
\]
The component forecast radii may be calibrated jointly, or by a Bonferroni split over value and the \(J\) constraints.
\end{assumption}


\begin{theorem}[Robust pacing certificate]
\label{thm:robust-certificate}
Let \(\widehat\pi\) be selected by Eq.~\eqref{eq:robust-policy}. Suppose Assumptions~\ref{ass:exchange}, \ref{ass:affine}, and \ref{ass:responsecoverage} hold, and suppose Eq.~\eqref{eq:robust-policy} is solved to optimization error \(\varepsilon_{\mathrm{opt}}\). Then, with probability at least \(1-\alpha-\alpha_\tau-\alpha_m\), all constraints hold for the realized future world:
\[
g_j(\widehat\pi;u^\star)\le 0,\qquad j=1,\ldots,J.
\]
Moreover, if \(\pi^\star_{\mathrm{rob}}\) is the best policy in \(\PiC\) satisfying the same robust constraints, then
\[
V(\pi^\star_{\mathrm{rob}};u^\star)-V(\widehat\pi;u^\star)
\le
2\sup_{\pi\in\PiC}
\left\{\widehat q^V_{1-\alpha_{\mathrm{comp}}}+\rho_V(\pi)\right\}
+\varepsilon_{\mathrm{opt}}.
\]
\end{theorem}

Theorem~\ref{thm:robust-certificate} ties the pieces to one pacing decision. A forecasting model helps when it reduces \(\widehat q^V\) and \(\widehat q^{g_j}\), regardless of whether it is simple or neural. Response measurement helps when it reduces \(\rho_V\) and any response-driven constraint radii. In the experiments, response uncertainty enters the value radius, while member-experience uncertainty enters the member-load constraint. A policy is credible only when its robust value clears the calibrated forecast, response, and experience margins. Otherwise the output is a shortlist rather than a forced recommendation.

The appendix adds three practical guardrails. Appendix Lemma~\ref{lem:quantile-concentration} underlies Proposition~\ref{prop:calibration-sample-size}. Appendix Theorem~\ref{thm:slack} shows why a point forecast can rank policies but cannot certify delivery, budget, or member-experience feasibility without slack in the relevant directions. Appendix Proposition~\ref{prop:catalog} justifies finite deployment catalogs. If the catalog is dense in the behavior that matters for value and constraints, optimizing over it loses only the corresponding approximation error.

\section{Experiments}

\subsection{Experimental Setup}

The empirical section evaluates pacing decisions coupled with forecast accuracy. Public production pacing logs with campaign budgets, delivery goals, member-load constraints, and randomized response are not available, so the experiments use public-data-calibrated streaming pacing benchmarks. Criteo supplies randomized ad response and anonymized context heterogeneity. KuaiRand supplies sequential recommendation exposure and repeated user-item structure. The pacing layer supplies the campaign catalog, delivery targets, budgets, member-load limits, and deployable policies. These benchmarks test whether the conformal pacing certificate behaves as predicted in reproducible environments.

Each case is loaded with up to \(500{,}000\) rows and converted into a \(120\)-block streaming planning instance with \(60\) training blocks, \(30\) calibration blocks, and \(30\) held-out test blocks. The main selector uses the medium catalog of \(12\) pacing policies (discussed in appendix Table \ref{tab:policy-catalog}), miscoverage levels \(\alpha=\alpha_\tau=\alpha_m=0.1\), component level \(\alpha_{\mathrm{comp}}=0.025\), and planning tolerance \(\eta_{\mathrm{plan}}=0.05\). The planning vector has coordinates for inventory mass, demand-pressure mass, and quality-adjusted opportunity mass within each segment. The selector is solved on the mean predicted held-out planning vector. Forecasts are generated by a rolling replay using time features and lagged block totals available at each block. Realized value and violation rates are computed by replaying the selected policy on each of the \(30\) held-out blocks. This matches the affine downstream-inventory structure used in the theory.

The streaming generator uses \(16\) context segments when available and \(6\) synthetic campaigns. Segment popularity, time-profile structure, quality profiles, base response, and treatment lift are calibrated from the public logs. Criteo uses pseudo-time induced by row order and a fixed conservative repeated-exposure scale, while KuaiRand uses date-hour-min structure and repeated user-item exposure. Campaign values, delivery targets, budgets, and member-load limits are then sampled from fixed seed-controlled ranges. Forecasting models are fit only on the \(60\) training blocks. The radii are calibrated only on the \(30\) calibration blocks and are never fit on the held-out test blocks. The forecasting diagnostic is not a leaderboard. It checks whether low prediction error and low pacing radius select the same upstream model. For reproducibility, the implemented comparison uses seven forecast generators: (i) previous-block point naive, (ii) seasonal ridge, (iii) nonlinear random features, (iv) histogram gradient boosting, (v) numpy-MLP (MultiLayer Perceptron), (vi) PyTorch-GRU (Gated Recurrent Unit), and (vii) PyTorch temporal Transformer. The GRU and Transformer models use automatic accelerator selection and run on CUDA when available.

\begin{table}[t]
\centering
\small
\caption{\small Empirical benchmark provenance. Criteo and KuaiRand provide public-data structure for response, context heterogeneity, time ordering, and repeated exposure, while the campaign budgets, delivery targets, member-load limits, pacing policies, and constraint system are constructed as a reproducible pacing layer. The experiments should therefore be considered as public-data-calibrated simulated stress tests, not as production pacing logs.}
\label{tab:benchmark-provenance}
\begin{tabularx}{\linewidth}{@{}>{\raggedright\arraybackslash}p{0.22\linewidth}YYY@{}}
\toprule
Component & Criteo source & KuaiRand source & Pacing layer \\
\midrule
Response/outcome signal & Randomized treatment and conversion & Logged random recommendation interactions & Calibrated value scale \\
Context heterogeneity & Anonymized feature buckets & Item and date-hour-min buckets & Context-to-campaign mapping \\
Time structure & Row-ordered pseudo-time bins & Date-hour-min sequence when available & \(120\)-block planning horizon \\
Repeated exposure/load & Limited in source log & Repeated user-item exposure & Member-load constraints \\
Campaign budgets and delivery goals & Not in public log & Not in public log & Constructed with fixed seed \\
Pacing catalog and policies & Not in public log & Not in public log & \(12\)-policy deployable catalog \\
\bottomrule
\end{tabularx}
\end{table}

\noindent \textbf{Dataset 1. Randomized ad response}:
The Criteo Uplift dataset provides randomized treatment assignment and ad-response outcomes \citep{diemert2018criteo}. We load up to \(250{,}000\) treated and \(250{,}000\) control rows when available, use the logged conversion outcome, construct \(64\) context identifiers from anonymized feature buckets, and create pseudo-time indices by ordering rows into chunks of size \(\lfloor n/96\rfloor\) before generating the streaming case. We use this dataset to estimate incremental response and advertiser value under different delivery intensities. Because the dataset is not a streaming ad server log, the benchmark constructs the pacing environment by ordering opportunities into time blocks and assigning campaigns to context clusters. Thus Criteo contributes randomized response and context heterogeneity, while the budgeted pacing problem is the reproducible stress layer.

\noindent \textbf{Dataset 2. Sequential recommendation exposure}:
KuaiRand provides repeated recommendation exposure with randomization and sequential user-item interactions \citep{gao2022kuairand}. We use the KuaiRand-Pure random log when the archive is available, load up to \(500{,}000\) rows, define the outcome as \(0.6\) times click plus \(0.3\) times long-view plus \(0.1\) times like, use date-hourmin bins for time, and construct \(64\) item-context buckets from video identifiers. We use it to calibrate opportunity arrival, repeated exposure, response decay, and load-sensitive experience costs. The campaign budgets, delivery constraints, and pacing policies remain constructed, as shown in Table~\ref{tab:benchmark-provenance}. When the loaded recommendation log does not contain enough treatment contrast for a stable response estimator, the implementation uses a conservative public-data-calibrated response scale and records this fallback in the response diagnostic.


The main experiments evaluate how calibrated uncertainty changes launch decisions and held-out constraint behavior. The first experiment compares point-forecast pacing, generic-residual conformal pacing, and the decision-calibrated selector under the same policy catalog; its results are reported in Figure~\ref{fig:main-pacing} and Table~\ref{tab:operational-replay-summary}. The second experiment reruns the selector with all seven forecasting models and asks whether the lowest held-out prediction error also gives the smallest calibrated radius in Eq.~\eqref{eq:dc-quantile}; its results are reported in Figure~\ref{fig:main-forecasters}. The third experiment compares predicted response, context-adjusted doubly robust response when treatment contrast is available, and robust response; its results are reported in Figure~\ref{fig:main-response}. The first and third main experiments use the numpy-MLP forecaster. The appendix reports the supporting diagnostics: (i) geometry and conformal coverage in Figures~\ref{fig:appendix-support-characterization}--\ref{fig:appendix-theory-geometry} and Table~\ref{tab:appendix-theory-checks}, (ii) catalog and deployment ablations in Table~\ref{tab:appendix-ablation-summary}, and (iii) calibration sample size in Figure~\ref{fig:appendix-calibration-size}.

\subsection{Main Results}

\begin{figure}[t]
  \centering
  \includegraphics[width=0.98\linewidth]{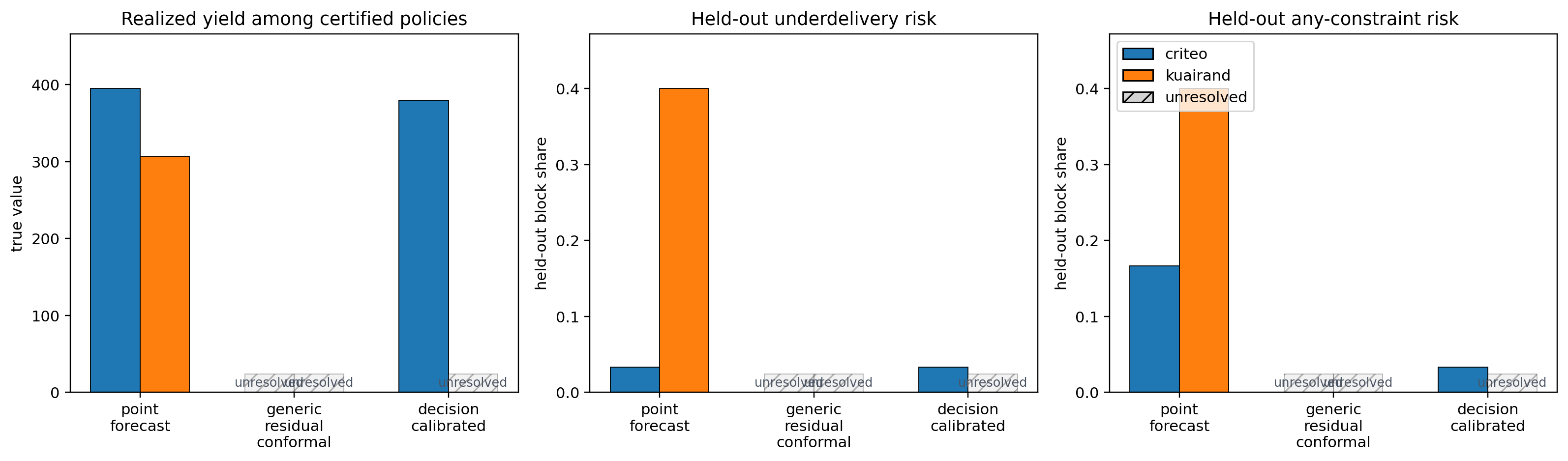}
  \caption{\small Deployment-facing replay comparison. Point-forecast pacing ignores forecast uncertainty. Generic-residual conformal pacing uses an unweighted residual radius, while the decision-calibrated selector uses component radii built from catalog sensitivities. Criteo has enough margin for a certified policy, while KuaiRand remains unresolved under repeated-exposure and member-load uncertainty.}
  \label{fig:main-pacing}
\end{figure}

\noindent \textbf{Decision-calibrated pacing versus point and generic robust baselines}:
Figure~\ref{fig:main-pacing} and Table~\ref{tab:operational-replay-summary} report the main selector comparison. The point-forecast baseline certifies aggressive policies without reserving forecast slack. On Criteo, point-forecast pacing selects \texttt{pace\_i3\_q0}, obtains realized value \(394.8\), and has a \(16.7\%\) any-violation rate over test blocks, including a \(13.3\%\) budget-violation rate. The generic-residual baseline uses the unweighted residual radius \(q_{\mathrm{generic}}=7236.7\), which is too conservative and returns unresolved. The decision-calibrated selector has Lagrangian radius \(q_{\mathrm{dc}}=18.4\) and uses component radii for value, delivery, budget, and member-load certification. It selects \texttt{pace\_i3\_q1}, obtains realized value \(379.5\), and reduces the any-violation rate to \(3.3\%\) with zero budget and member-load violations in the realized test blocks.

The KuaiRand case is harder because repeated exposure creates larger delivery and member-load uncertainty. Point-forecast pacing selects \texttt{pace\_i3\_q2} with a \(40.0\%\) delivery-driven violation rate. The generic-residual baseline again returns unresolved. The decision-calibrated radius falls from \(4629.4\) to \(278.6\), but the selector still returns unresolved after the value, delivery, budget, and member-load margins are applied. This is the intended behavior. The radius can be much smaller than a generic residual band, while certification still fails when constraint margins are weak.


Appendix~\ref{app:empirical-diagnostics} explains why the Criteo certification occurs and why the KuaiRand case remains unresolved. The decomposition is summarized in Table~\ref{tab:appendix-ablation-summary}. First, holding the robust optimizer fixed, replacing the generic residual radius with the decision-calibrated component radii changes the Criteo robust selector from unresolved to certified. This isolates the role of catalog sensitivity geometry in the sense that the generic radius is too large because it pays for forecast-error directions that the policy catalog does not use. Second, holding the decision-calibrated certificate fixed, the nominal point-forecast policy \texttt{pace\_i3\_q0} fails the robust budget check, with robust overspend \(34.8\). The robust selector instead moves to \texttt{pace\_i3\_q1}, whose robust budget margin is \(-360.7\). This isolates the role of robust optimization in the sense that once uncertainty is measured in the right directions, the optimizer still has to choose a policy with enough budget and member-load slack. On KuaiRand, the decision-calibrated radius is much smaller than the generic residual radius, but all robust margins remain weak, so the correct output remains unresolved.

\begin{table}[t]
\centering
\scriptsize
\caption{\small Deployment-facing replay summary. The radius column reports the scalar headline radius, with \(0\) for point forecasts, \(q_{\mathrm{generic}}\) for generic conformal pacing, and \(q_{\mathrm{dc}}\) for the decision-calibrated selector. Certification additionally uses separate component radii for value, delivery, budget, and member load. Unresolved rows have no launched policy, so realized value and held-out violation rates are not reported.}
\label{tab:operational-replay-summary}
\begin{tabular}{@{}lllp{0.13\linewidth}rrrrr@{}}
\toprule
Case & Method & Decision & Policy & Shown radius & Value & Any viol. & Budget viol. & Member viol. \\
\midrule
Criteo & Point forecast & certify & \texttt{pace\_i3\_q0} & \(0.0\) & \(394.8\) & \(16.7\%\) & \(13.3\%\) & \(0.0\%\) \\
Criteo & Generic conformal & unresolved & -- & \(7236.7\) & -- & -- & -- & -- \\
Criteo & Decision calibrated & certify & \texttt{pace\_i3\_q1} & \(18.4\) & \(379.5\) & \(3.3\%\) & \(0.0\%\) & \(0.0\%\) \\
KuaiRand & Point forecast & certify & \texttt{pace\_i3\_q2} & \(0.0\) & \(306.9\) & \(40.0\%\) & \(0.0\%\) & \(0.0\%\) \\
KuaiRand & Generic conformal & unresolved & -- & \(4629.4\) & -- & -- & -- & -- \\
KuaiRand & Decision calibrated & unresolved & -- & \(278.6\) & -- & -- & -- & -- \\
\bottomrule
\end{tabular}
\end{table}

\begin{figure}[t]
  \centering
  \includegraphics[width=0.98\linewidth]{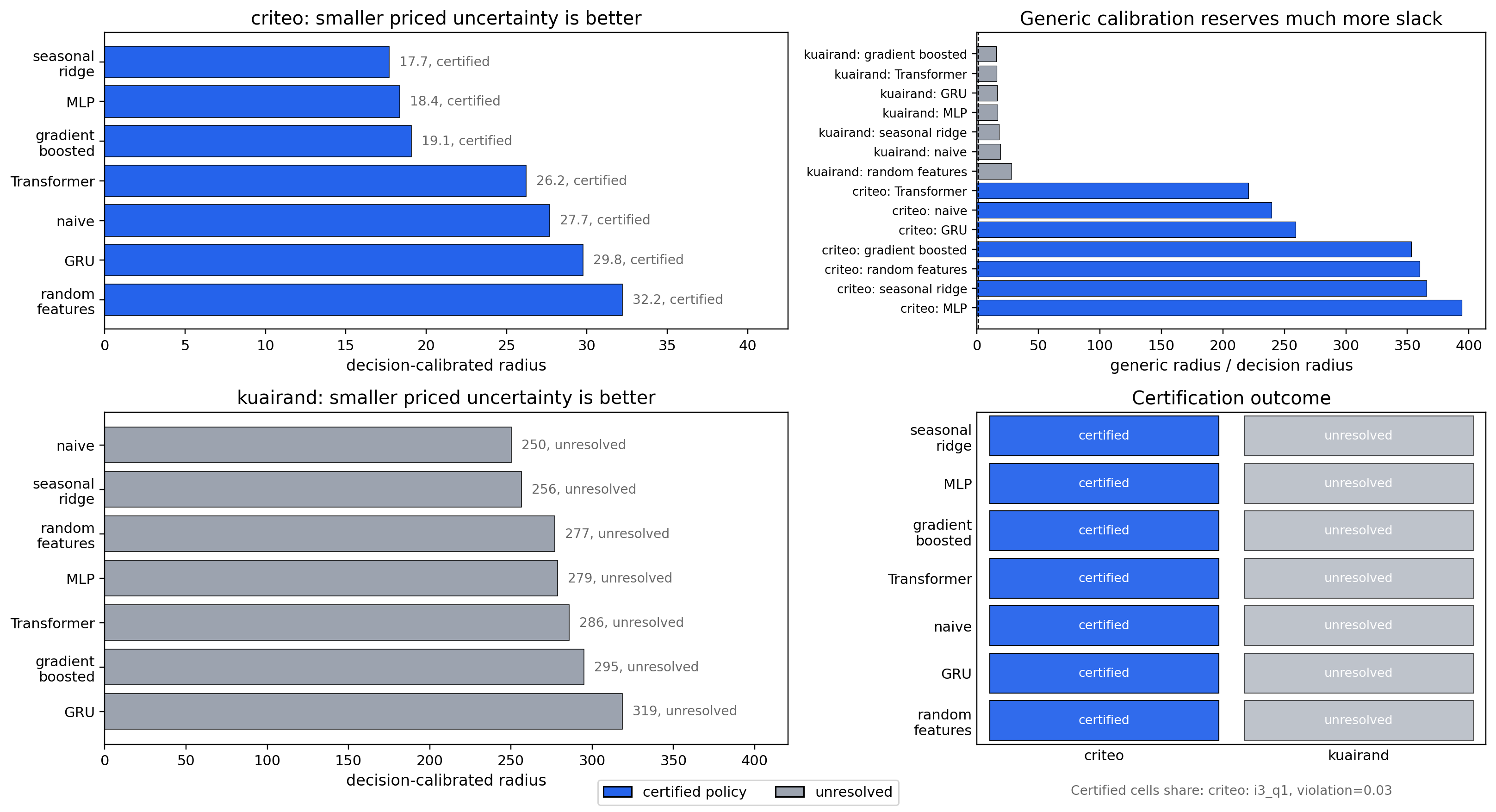}
 \caption{\small Forecasting inputs evaluated by downstream pacing uncertainty and certification outcome. The top-left panel reports Criteo decision-calibrated radii by forecaster; smaller radii mean less uncertainty in the value and constraint directions used by the pacing selector, and all Criteo forecasters certify a policy. The bottom-left panel reports the same quantity for KuaiRand, where every forecaster remains unresolved despite substantially smaller radii than generic residual calibration. The top-right panel shows the ratio between the generic residual radius and the decision-calibrated radius; large ratios indicate how much extra slack generic calibration reserves for forecast-error directions that are not priced by the policy catalog. The bottom-right panel summarizes the final decision state by dataset and forecaster, with blue cells denoting certified policies and gray cells denoting unresolved decisions. The figure shows that forecast accuracy alone is not the relevant object for pacing. The useful quantity is the calibrated radius after forecast errors are projected through the value and constraint sensitivities of the deployable policy catalog.}
  \label{fig:main-forecasters}
\end{figure}

\noindent \textbf{Forecast accuracy is not the same as decision uncertainty}:
Figure~\ref{fig:main-forecasters} is not a forecasting leaderboard. Its point is that the lowest held-out error need not produce the smallest pacing radius. On Criteo, the Transformer has the lowest test MAE among the fitted models (\(481.0\)), but the smallest decision radius is obtained by the seasonal ridge model (\(17.7\)), followed closely by the numpy MLP (\(18.4\)) and gradient-boosted model (\(19.1\)). All Criteo forecasters certify \texttt{pace\_i3\_q1} with realized value \(379.5\). On KuaiRand, the point-naive forecaster has the smallest decision radius (\(250.1\)), but every forecaster remains unresolved because member-load and delivery uncertainty dominate the robust constraints. A forecasting model is useful for pacing when it shrinks uncertainty in directions used by the pacing problem, not merely when it improves generic prediction.

\begin{figure}[t]
  \centering
  \includegraphics[width=0.95\linewidth]{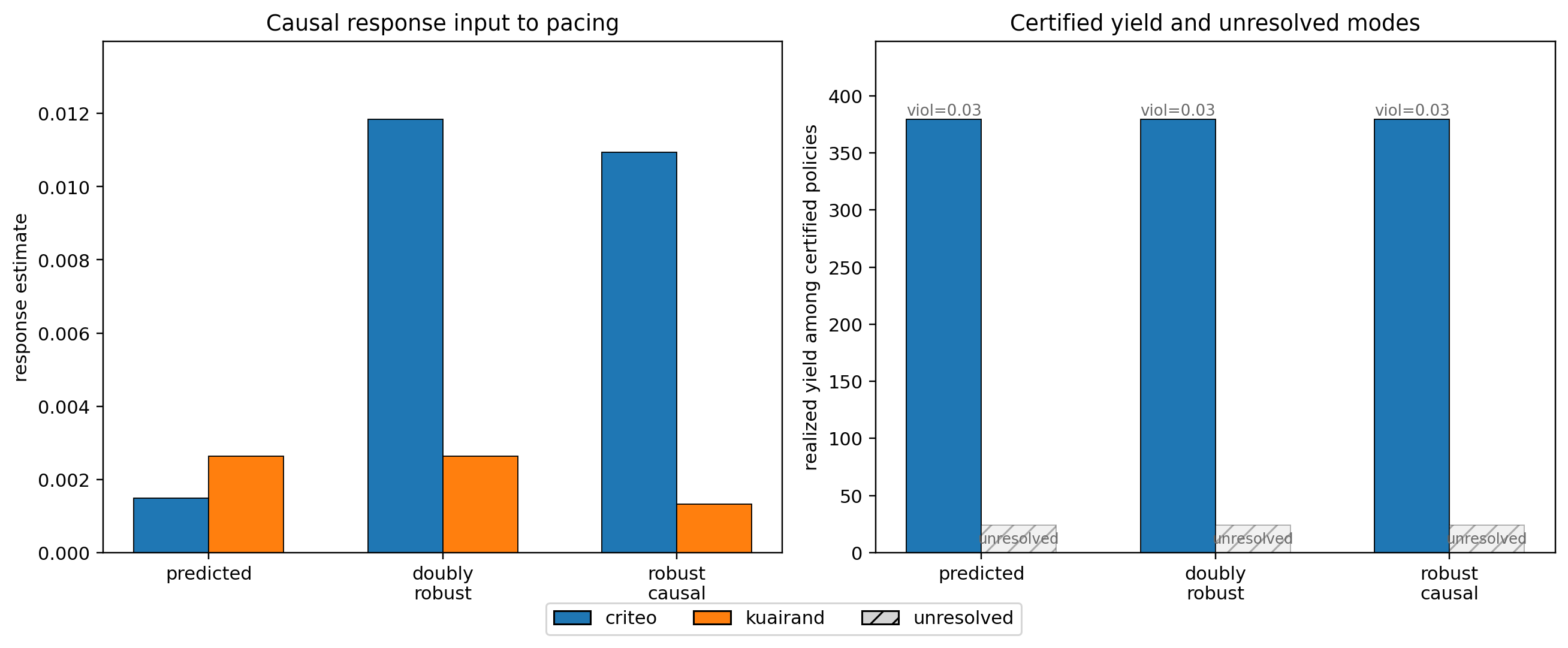}
    \caption{\small Response-estimation inputs to the conformal pacing selector. The left panel compares the response scale supplied to the pacing problem under three modes: (i) a model-predicted response, (ii) a context-adjusted doubly robust estimate when treatment contrast is available, and (iii) a robust response input that subtracts the response uncertainty radius. Criteo has enough randomized treatment contrast for the doubly robust estimate, so its response input increases relative to the raw prediction and remains positive after robust adjustment. KuaiRand has weaker treatment contrast in the loaded recommendation log, so the implementation uses a conservative public-data-calibrated response fallback and records the uncertainty in the robust certificate. The right panel shows the downstream pacing outcome under the same three response modes. Criteo remains certified in all modes, with realized value around \(379.5\) and held-out any-violation rate \(0.03\). KuaiRand remains unresolved in all modes because repeated-exposure and member-load uncertainty keep the robust constraints from clearing, even when the response input changes.}
  \label{fig:main-response}
\end{figure}

\noindent \textbf{Response uncertainty enters the same pacing certificate.}
Figure~\ref{fig:main-response} checks how the incremental-response input affects the selector. The implementation first computes a public-data response summary. When both treatment arms have usable support, it forms a context-adjusted doubly robust score. Treatment propensities are estimated within context buckets and clipped to \([0.05,0.95]\), treated and control outcome regressions are replaced by context-level means, and the response estimate is the average doubly robust score. Its standard error is the empirical standard deviation of the doubly robust scores divided by the square root of the number of rows. The robust response input is the lower confidence adjustment
$
\widehat\tau_{\mathrm{rob}}
=
\max\{\widehat\tau_{\mathrm{DR}}-1.96\,\widehat{\mathrm{se}}_{\mathrm{DR}},10^{-4}\}.
$
When the loaded log has insufficient treatment variation, the implementation records this fallback and uses a conservative public-data-calibrated response scale.

For Criteo, the raw predicted response estimate is \(0.00149\), the context-adjusted doubly robust estimate is \(0.01184\) with standard error \(0.00046\), and the robust response input is \(0.01094\). All three response modes select \texttt{pace\_i3\_q1}. Robust value changes because the response scale changes, but the certified policy is stable. For KuaiRand, the loaded recommendation log does not contain enough treatment contrast for a stable doubly robust estimator. The experiment therefore uses the conservative fallback response scale, carries the response standard error into \(\rho_V(\pi)\), and remains unresolved under predicted, doubly robust, and robust response modes. Thus response estimation is treated as an input to the same robust certificate as forecast and member-experience uncertainty, rather than as a separate decision rule.

\noindent \textbf{Summary}:
Across the main experiments, the decision-calibrated radii change both the radius and the decision state. On Criteo, they certify a policy that is less risky than the point-forecast policy and far less conservative than generic-residual conformal pacing. On KuaiRand, they reduce the generic radius but still return unresolved because repeated-exposure and member-experience uncertainty leave no robust certificate. Appendix~\ref{app:empirical-diagnostics} reports the supporting diagnostics and ablations.

\FloatBarrier

\section{Conclusion}

We develop a conformal uncertainty framework for pacing decisions in streaming advertising. The framework weights forecast residuals by objective and constraint sensitivity directions, calibrates the resulting radius, combines it with response and member-experience uncertainty, and selects implementable policies by robust constrained optimization. A policy is selected when it improves robust value while satisfying delivery and experience constraints. When uncertainty is too large for a single recommendation, the selector returns a shortlist.

\bibliographystyle{plainnat}
\bibliography{references}

\appendix

\section{Supporting Results}

The appendix has three roles. Appendix~A and C contains proof details and supporting theory. Appendix~B first reports theory-supporting diagnostics for the support-function geometry, conformal coverage, robust certificate, slack condition, and finite catalog approximation. It then reports deployment-facing ablations for catalog granularity, response scale, budget pressure, member-load pressure, and catalog geometry versus robust optimization. The final diagnostic studies calibration sample size. Readers mainly interested in the method's empirical behavior can start with Appendix~B. Readers checking the formal guarantees can start with the proofs in Appendix~A and C.

The quantile lemma is the sampling fact that turns conformal validity into a planning calculation. Split conformal prediction gives coverage for any exchangeable calibration set. The lemma below asks how many calibration blocks are needed for the empirical radius to be close to the population radius that would be used with unlimited calibration data.

\begin{lemma}[Uniform quantile concentration for calibrated decision scores]
\label{lem:quantile-concentration}
Let \(S_1,\ldots,S_n\) be exchangeable calibration scores with distribution function \(F\), and let
\[
q=F^{-1}(1-\alpha).
\]
Assume \(F\) has density at least \(f_{\min}>0\) on
\[
[q-\varepsilon,\ q+\varepsilon],
\]
with \(f_{\min}\varepsilon\le 1\). Let \(\widehat q\) be the split-conformal empirical quantile
\[
\widehat q
=
S_{(\lceil (n+1)(1-\alpha)\rceil)},
\]
where \(S_{(r)}\) denotes the \(r\)th order statistic. There is a universal constant \(C>0\) such that, for any \(\delta\in(0,1)\), if
\[
n
\ge
\frac{C}{f_{\min}^2\varepsilon^2}
\log\frac{2}{\delta},
\]
then
\[
\Pbb\left(|\widehat q-q|\le \varepsilon\right)
\ge
1-\delta.
\]
\end{lemma}

The slack result formalizes a common failure mode. A point forecast can make a candidate policy look feasible because its nominal constraint value is negative. If the forecast can move in a direction that increases the same constraint, nominal feasibility is not a deployment guarantee. The relevant margin is the largest plausible movement in the constraint's sensitivity direction.

\begin{theorem}[Necessity of decision-calibrated slack]
\label{thm:slack}
Consider any policy \(\pi\) and constraint \(j\) whose inventory contribution is affine with sensitivity vector \(a_{\pi,j}\). Suppose the plausible forecast-error set contains both perturbations \(e\) and \(-e\) with \(|\langle a_{\pi,j},e\rangle|=\delta\). If the point-forecast margin satisfies
\[
-\widehat g_j(\pi)<\delta,
\]
then there exists a plausible future in the same decision-calibrated uncertainty set for which \(g_j(\pi)>0\). Consequently, no point-forecast pacing rule can certify constraint \(j\) over this uncertainty class unless it keeps slack at least \(\delta\) in the constraint's decision-relevant direction.
\end{theorem}

If the nominal margin is smaller than the uncertainty in the constraint direction, there is an admissible future in which the policy violates the constraint. In ads language, a plan that appears to satisfy delivery, budget, or ad-load limits under the point forecast can still fail when scarce inventory moves in the direction that affects that constraint. Algorithm~\ref{alg:selector} therefore certifies policies only after subtracting forecast, response, and member-experience slack.

The following finite-catalog result addresses a different implementation issue. In principle, a pacing policy could be any measurable mapping from histories and contexts to bid multipliers or allocation probabilities. In production, platforms usually deploy an auditable catalog of pacing rules, budget multipliers, reserve schedules, or load-control policies. The proposition gives the condition under which this restriction is harmless at the resolution relevant for robust pacing.

\begin{proposition}[Finite catalog sufficiency in pacing geometry]
\label{prop:catalog}
Let \(R(\pi)\) be the robust pacing risk associated with Eq.~\eqref{eq:robust-policy}, and let \(\U\) be the uncertainty set of plausible future worlds. Suppose \(R\) is \(L_\Pi\)-Lipschitz under the pacing-behavior metric
\[
d_{\mathrm{pace}}(\pi,\pi')
=
\sup_{u\in\U}
\left(
|V(\pi;u)-V(\pi';u)|
+\sum_{j=1}^J |g_j(\pi;u)-g_j(\pi';u)|
\right).
\]
If \(\PiC\) is an \(\eta\)-net of a richer policy class \(\Pi\) under \(d_{\mathrm{pace}}\), then
\[
\inf_{\pi\in\PiC}R(\pi)
\le
\inf_{\pi\in\Pi}R(\pi)+L_\Pi\eta.
\]
\end{proposition}

The metric \(d_{\mathrm{pace}}\) compares policies by robust value and constraint behavior across plausible futures. The catalog need not approximate every microscopic action probability. It only needs to approximate value, delivery, budget, and experience behavior. If the spacing is \(\eta\), deployment discreteness costs at most \(L_\Pi\eta\). This justifies the catalog diagnostics. Adding pacing rules is useful only when it materially changes robust value or constraint behavior under \(d_{\mathrm{pace}}\).

\section{Empirical Diagnostics and Ablations}
\label{app:empirical-diagnostics}

This appendix records diagnostics that support the main empirical story. The main body focuses on three deployment-facing comparisons: (i) pacing method, (ii) forecasting radius, and (iii) response mode. The diagnostics below isolate the selector components and stress-test the pipeline under catalog, response, budget, member-load, and calibration-size perturbations.

\subsection{Geometry, conformal coverage, and certificate checks}

\begin{figure}[t]
  \centering
  \includegraphics[width=0.95\linewidth]{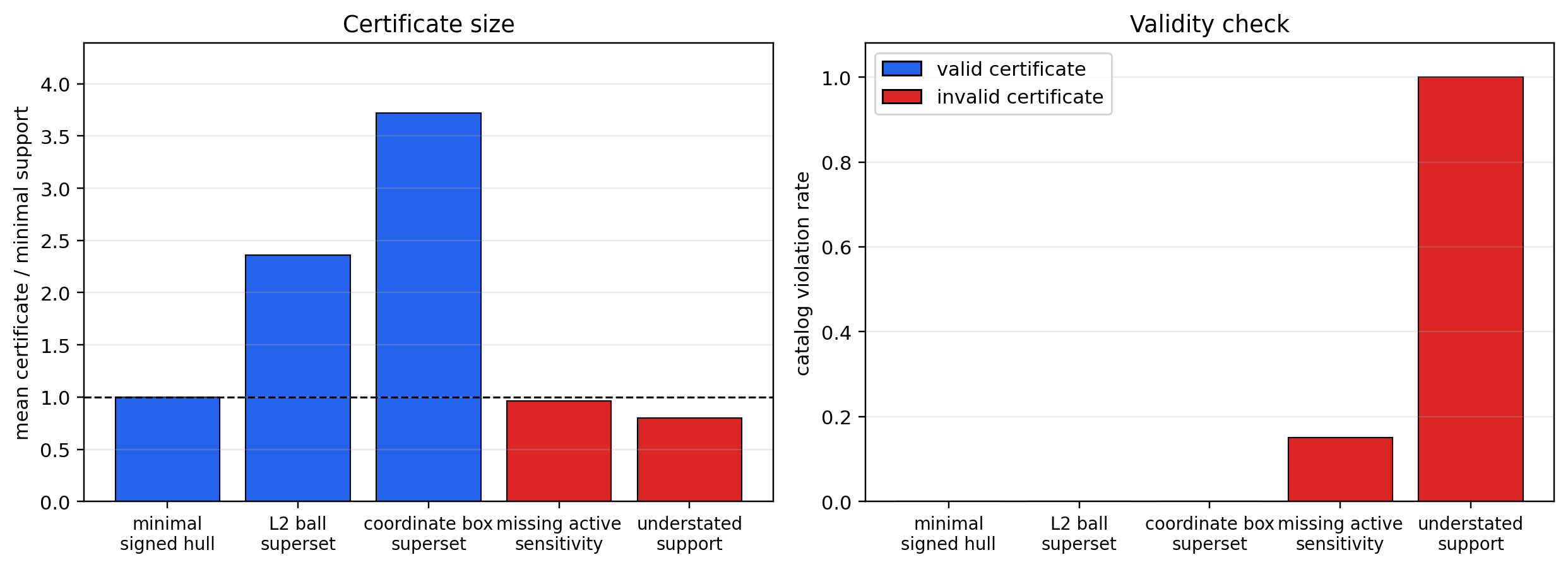}
    \caption{\small Geometry diagnostic for Theorem~\ref{thm:sharp-geometry}. This is a synthetic theorem-level diagnostic rather than a Criteo or KuaiRand experiment. The implementation constructs \(8\) orthogonal policy-sensitivity directions in a \(16\)-dimensional forecast-error space and evaluates \(500\) random forecast-error draws plus targeted active-direction errors. The left panel reports each candidate certificate's mean size relative to the minimal signed-hull support function. The right panel reports the catalog violation rate, meaning the share of tested forecast errors for which the candidate certificate is smaller than the true catalog support score. Supersets of the signed sensitivity hull are valid but conservative. Certificates that omit an active signed sensitivity or scale below the support function violate catalog validity.}
  \label{fig:appendix-support-characterization}
\end{figure}

\begin{figure}[t]
  \centering
  \includegraphics[width=0.85\linewidth]{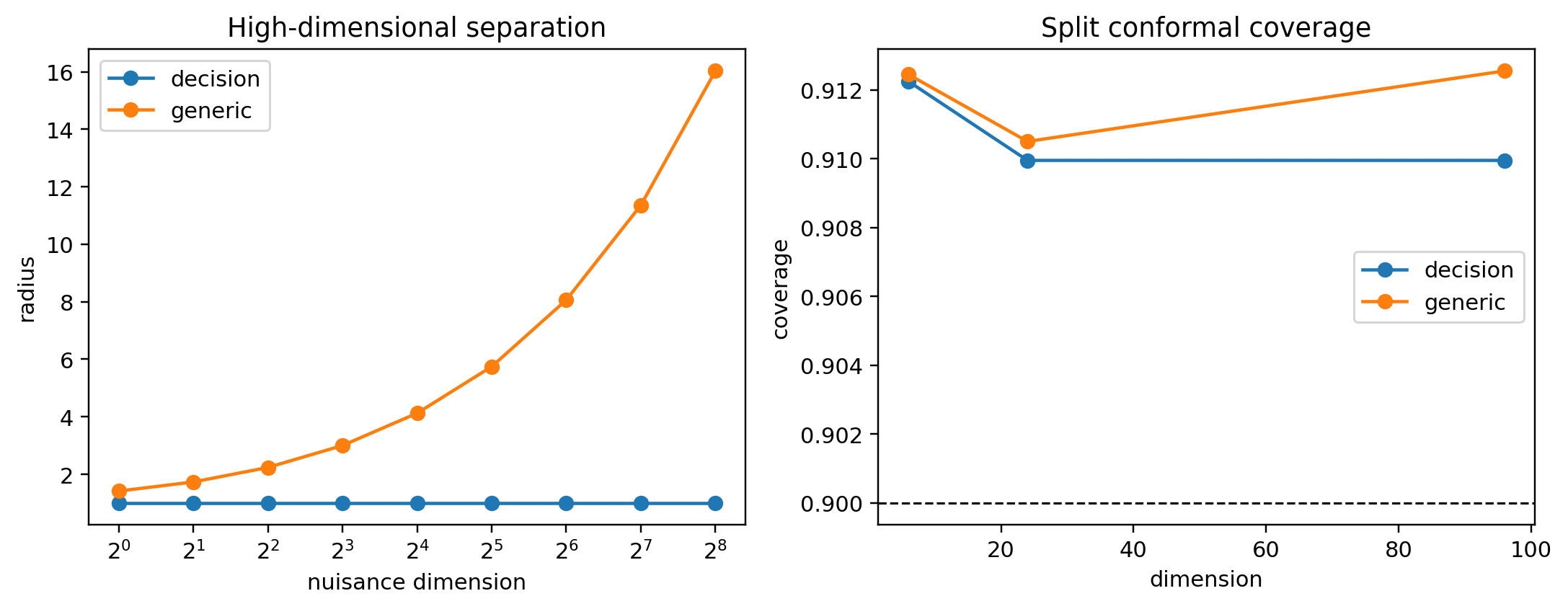}
  \caption{\small Theory-level separation and coverage diagnostics. These are synthetic diagnostics for Theorem~\ref{thm:separation} and the split-conformal coverage statement in Theorem~\ref{thm:sharp-geometry}. The left panel constructs forecast-error vectors with one decision-relevant coordinate and an increasing number of nuisance inventory coordinates that are orthogonal to every policy-sensitivity direction. The decision-calibrated radius stays fixed at \(1\), because the nuisance coordinates do not affect any deployable policy in the catalog. The generic Euclidean residual radius grows from about \(1.41\) to \(16.03\) as the nuisance dimension increases to \(256\), showing the separation predicted by Theorem~\ref{thm:separation}. The right panel verifies split-conformal coverage over synthetic Gaussian forecast-error draws for dimensions \(6\), \(24\), and \(96\). The dashed horizontal line is the target coverage \(0.90\). Both the decision-calibrated score and the generic residual score achieve coverage slightly above the target, while only the decision-calibrated score avoids paying for nuisance dimensions.}
  \label{fig:appendix-theory-geometry}
\end{figure}

Figures~\ref{fig:appendix-support-characterization} and \ref{fig:appendix-theory-geometry} isolate the geometry theorem. The equality check has maximum absolute gap \(0\). Figure~\ref{fig:appendix-support-characterization} compares five scalar certificates on the same policy-sensitivity set. The minimal signed hull has mean certificate ratio \(1.00\) and no violations. Two strict supersets, an enclosing \(\ell_2\) ball and a coordinate box, are valid but larger, with mean ratios \(2.36\) and \(3.72\). A set that omits one active signed sensitivity violates the catalog certificate on \(15.1\%\) of tested forecast-error directions, and the understated radius \(0.8\Phi_{\PiC}\) fails on every tested draw. This matches Theorem~\ref{thm:sharp-geometry}. Valid certificates contain the signed policy sensitivities, and the signed hull is the minimal such set. Figure~\ref{fig:appendix-theory-geometry} verifies split-conformal coverage and high-dimensional separation. For dimensions \(6\), \(24\), and \(96\), the decision-calibrated coverage rates are \(0.912\), \(0.910\), and \(0.910\), respectively, against a target of \(0.900\). The separation panel shows the generic residual radius increasing from \(1.41\) to \(16.03\) as the nuisance dimension grows to \(256\), while the decision-calibrated radius remains fixed at \(1\).

Table~\ref{tab:appendix-theory-checks} summarizes the remaining theory diagnostics. Their goal is not to mimic Criteo or KuaiRand, but to check that each mathematical component behaves as the theory predicts. The robust-certificate diagnostic tests Theorem~\ref{thm:robust-certificate}. It simulates separate uncertainty layers for the inventory forecast, incremental response, and member-experience model, then runs the same robust-selection logic as Algorithm~\ref{alg:selector}. With \(\alpha=\alpha_\tau=\alpha_m=0.1\), the theorem gives a conservative joint coverage target of
$
1-\alpha-\alpha_\tau-\alpha_m=0.700.
$
Across the tested dimensions, the observed joint coverage rates are \(0.718\), \(0.732\), and \(0.718\). The realized certificate violation rate is at most \(0.002\), and the regret bound in Theorem~\ref{thm:robust-certificate} holds in at least \(0.998\) of simulated runs. This confirms that the union-bound certificate is conservative but operationally valid in the constructed affine setting.

The slack diagnostic tests Appendix Theorem~\ref{thm:slack}. It constructs an affine constraint with directional uncertainty size \(\delta=0.2\). When the point-forecast margin is smaller than \(\delta\), an admissible forecast perturbation can push the realized constraint above zero, so the nominally feasible policy is not certifiable. When the margin is at least \(\delta\), the same constructed uncertainty set is safely absorbed by the slack. This illustrates why Algorithm~\ref{alg:selector} certifies policies only after tightening constraints by the calibrated radii.

The finite-catalog diagnostic tests Appendix Proposition~\ref{prop:catalog}. It compares increasingly fine catalogs, with sizes from \(4\) to \(40\), against the Lipschitz-net approximation bound. In every tested setting, the observed catalog approximation error remains below the bound. This supports the use of a finite deployable catalog: the catalog need not approximate every possible pacing rule, only the value and constraint behavior relevant to robust selection.


\begin{table}[t]
\centering
\small
\caption{\small Appendix diagnostics associated with the theoretical results.}
\label{tab:appendix-theory-checks}
\begin{tabularx}{\linewidth}{@{}>{\centering\arraybackslash}p{0.07\linewidth}>{\raggedright\arraybackslash}p{0.22\linewidth}>{\raggedright\arraybackslash}p{0.30\linewidth}X@{}}
\toprule
S.No. & Result component & Implementation diagnostic & Observed outcome \\
\midrule
1 & Sharp support function & Equality and understated-radius check & Gap \(0\), and \(0.8\Phi_{\PiC}\) fails on all draws \\
2 & Signed-hull characterization & Minimal hull, supersets, missing active sensitivity & Supersets valid but larger, and missing sensitivity fails on \(15.1\%\) \\
3 & Split-conformal coverage & Coverage over dimensions \(6,24,96\) & \(0.912,0.910,0.910\) at target \(0.900\) \\
4 & High-dimensional separation & Nuisance dimension \(1\) to \(256\) & Generic/decision ratio grows to \(16.03\) \\
5 & Robust certificate & Simulated selector coverage and regret & Joint coverage \(0.718\)--\(0.732\) at target \(0.700\), violation \(\le0.002\), and regret bound holds \(\ge0.998\) \\
6 & Slack necessity & Margin versus \(\delta=0.2\) & Margins below \(\delta\) can violate \\
7 & Finite catalog & Catalog sizes \(4\) to \(40\) & Approximation error below \(L_\Pi\eta\) \\
\bottomrule
\end{tabularx}
\end{table}

\subsection{Policy catalog used in the main experiments}

Table~\ref{tab:policy-catalog} reports the \(12\)-policy medium catalog used in the main Criteo and KuaiRand experiments. The policy names are implementation labels. The intensity parameter controls how aggressively the policy uses available opportunities. The quality-focus parameter shifts allocation toward quality-adjusted opportunity mass. The demand-focus, load-guard, and budget-guard parameters are deterministic functions used by the implementation:
$
\mathrm{demand\_focus}=1-0.5\,\mathrm{quality\_focus},
$ $
\mathrm{load\_guard}=0.8+0.4\,\mathrm{quality\_focus},
$ $
\mathrm{budget\_guard}=1.15-0.25\,\mathrm{intensity}.
$
Thus higher quality focus gives more protection against member-load pressure, while higher intensity increases delivery and value potential but also increases budget pressure.

\begin{table}[t]
\centering
\small
\caption{\small Medium pacing-policy catalog used in the main experiments. The catalog crosses four intensity levels with three quality-focus levels, giving \(12\) deployable policies.}
\label{tab:policy-catalog}
\begin{tabular}{@{}lrrrrr@{}}
\toprule
Policy & Intensity & Quality focus & Demand focus & Load guard & Budget guard \\
\midrule
\texttt{pace\_i0\_q0} & \(0.40\) & \(0.15\) & \(0.925\) & \(0.86\) & \(1.05\) \\
\texttt{pace\_i0\_q1} & \(0.40\) & \(0.60\) & \(0.700\) & \(1.04\) & \(1.05\) \\
\texttt{pace\_i0\_q2} & \(0.40\) & \(0.95\) & \(0.525\) & \(1.18\) & \(1.05\) \\
\texttt{pace\_i1\_q0} & \(0.60\) & \(0.15\) & \(0.925\) & \(0.86\) & \(1.00\) \\
\texttt{pace\_i1\_q1} & \(0.60\) & \(0.60\) & \(0.700\) & \(1.04\) & \(1.00\) \\
\texttt{pace\_i1\_q2} & \(0.60\) & \(0.95\) & \(0.525\) & \(1.18\) & \(1.00\) \\
\texttt{pace\_i2\_q0} & \(0.80\) & \(0.15\) & \(0.925\) & \(0.86\) & \(0.95\) \\
\texttt{pace\_i2\_q1} & \(0.80\) & \(0.60\) & \(0.700\) & \(1.04\) & \(0.95\) \\
\texttt{pace\_i2\_q2} & \(0.80\) & \(0.95\) & \(0.525\) & \(1.18\) & \(0.95\) \\
\texttt{pace\_i3\_q0} & \(1.00\) & \(0.15\) & \(0.925\) & \(0.86\) & \(0.90\) \\
\texttt{pace\_i3\_q1} & \(1.00\) & \(0.60\) & \(0.700\) & \(1.04\) & \(0.90\) \\
\texttt{pace\_i3\_q2} & \(1.00\) & \(0.95\) & \(0.525\) & \(1.18\) & \(0.90\) \\
\bottomrule
\end{tabular}
\end{table}

\subsection{Sensitivity ablations}

Table~\ref{tab:appendix-ablation-summary} summarizes the sensitivity ablations that support the main empirical results in Figure~\ref{fig:main-pacing}, Table~\ref{tab:operational-replay-summary}, Figure~\ref{fig:main-forecasters}, and Figure~\ref{fig:main-response}. These diagnostics ask whether the conclusions depend on one catalog, one response scale, one budget or member-load setting, or one way of measuring forecast error.

The catalog-granularity diagnostic tests Appendix Proposition~\ref{prop:catalog}. In Criteo, the selected policy changes as the catalog is refined. The coarse catalog selects \texttt{pace\_i2\_q0} with value \(393.1\), the medium catalog used in the main experiments selects \texttt{pace\_i3\_q1} with value \(379.5\), and the fine catalog selects \texttt{pace\_i5\_q2} with value \(413.7\). All three have \(3.3\%\) any-violation rate in the realized test blocks. This shows why the finite catalog matters as  refinement is useful when it changes the robust value and constraint behavior available to the selector.

The response misspecification diagnostic complements Figure~\ref{fig:main-response}. It perturbs the assumed Criteo response scale from \(0.5\) to \(2.0\) times the calibrated value and compares no-response-radius and robust-response-radius modes. Across all twelve settings, the selector returns \texttt{pace\_i3\_q1}. This stability should be interpreted narrowly. The selected Criteo policy is robust to this response-scale stress grid, not to arbitrary response uncertainty.

The KuaiRand budget/member-load diagnostic complements the unresolved KuaiRand result in Figure~\ref{fig:main-pacing} and Table~\ref{tab:operational-replay-summary}. It multiplies the budget limits by \(0.7,1.0,1.4\) and the member-load limits by \(0.75,1.0,1.25\). The implementation records the best unresolved representative policy and its robust delivery, budget, and member-load margins in each setting. All nine settings remain unresolved, matching the main result that repeated-exposure and member-load uncertainty dominate certification in that case.

The catalog-geometry-versus-robustness ablation decomposes the main Criteo gain in Figure~\ref{fig:main-pacing}. It uses four rows for each case: a nominal point-forecast selector with zero radii, a generic-residual robust selector, a decision-calibrated robust selector, and an audit of the nominal policy under the same decision-calibrated certificate. The generic and decision-calibrated rows use the same robust optimizer, so their difference isolates the uncertainty geometry. The audit row holds the decision-calibrated certificate fixed and asks whether the nominal policy would have passed it, isolating robust optimization. On Criteo, catalog geometry reduces the robust radius from \(7236.7\) to \(18.4\) and changes the robust selector from unresolved to certified. The nominal policy \texttt{pace\_i3\_q0} has higher realized value but fails the certificate because its robust budget overspend is \(34.8\). The robust selector chooses \texttt{pace\_i3\_q1}, whose robust budget margin is \(-360.7\) and whose realized any-violation rate is \(3.3\%\) rather than \(16.7\%\). On KuaiRand, the radius falls from \(4629.4\) to \(278.6\), but both robust selectors remain unresolved and the nominal policy fails all robust margins.

The dual-versus-unweighted diagnostic complements Figure~\ref{fig:main-forecasters}. On Criteo, the unweighted residual radius is \(220.6\) to \(394.2\) times larger than the dual-weighted Lagrangian radius across the fitted forecasters. This supports the central claim that generic forecast calibration can reserve slack for inventory directions that no deployable policy uses, while the decision-calibrated radius pays only for relevant directions.

\begin{table}[t]
\centering
\small
\caption{\small Sensitivity-ablation summary. Values are read from the saved appendix artifacts.}
\label{tab:appendix-ablation-summary}
\begin{tabularx}{\linewidth}{@{}>{\raggedright\arraybackslash}p{0.20\linewidth}YY@{}}
\toprule
Ablation & Stress grid & Main observation \\
\midrule
Catalog granularity & Coarse, medium, fine Criteo catalogs & Fine catalog reaches value \(413.7\) with \(3.3\%\) violation rate \\
Response scale & Multipliers \(0.5\) to \(2.0\) & Criteo remains at \texttt{pace\_i3\_q1} in all settings \\
Budget/member pressure & \(3\times3\) KuaiRand stress grid & All settings unresolved, with the saved table reporting representative robust margins \\
Catalog geometry vs. robustness & Point, generic-robust, decision-calibrated robust, nominal-policy audit & Criteo certification comes from geometry, and robust optimization rejects the nominal policy's budget-unsafe certificate \\
Dual versus unweighted & Seven Criteo forecast generators & Generic radius is \(220.6\)--\(394.2\times\) larger \\
\bottomrule
\end{tabularx}
\end{table}

\subsection{Calibration sample size}

\begin{figure}[t]
  \centering
  \includegraphics[width=0.98\linewidth]{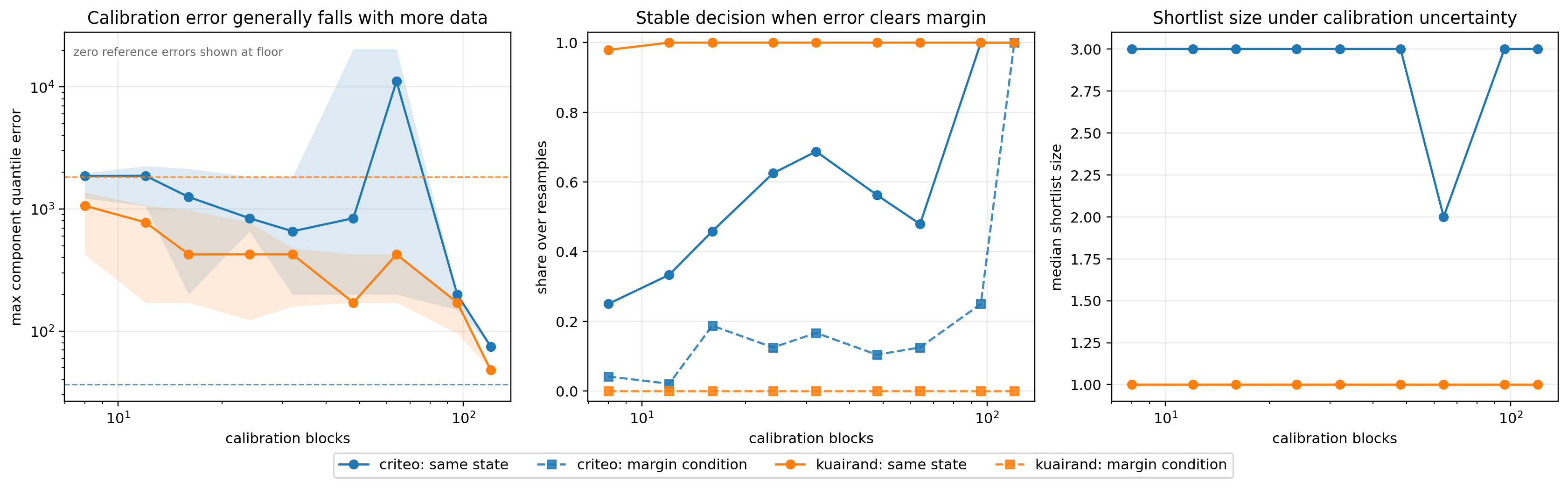}
  \caption{\small Calibration sample-size diagnostic for Proposition~\ref{prop:calibration-sample-size}. The experiment builds longer \(480\)-block Criteo and KuaiRand streaming cases, treats the full \(120\)-block calibration split as the reference calibration set, then repeatedly subsamples smaller calibration sets and reruns the same component-radius calibration and robust selector. The left panel reports the median maximum component quantile error across value, delivery, budget, and member-load radii, relative to the full-calibration reference. Shaded bands show the interquartile range over resamples, and dashed horizontal lines show half of the empirical decision margin for each case. Because the vertical axis is logarithmic, exact zero errors relative to the reference are displayed at a small positive floor. The middle panel reports two stability rates: (i) the solid line is the share of resamples recovering the same decision state as the full-calibration reference, and (ii) the dashed line is the share satisfying the sufficient margin condition from Proposition~\ref{prop:calibration-sample-size}. The right panel reports the median shortlist size under calibration uncertainty. Criteo stabilizes to the reference certified policy as calibration size grows, while KuaiRand stabilizes to the unresolved state because member-load and delivery margins remain weak even with small calibration error.}
  \label{fig:appendix-calibration-size}
\end{figure}

Figure~\ref{fig:appendix-calibration-size} tests Proposition~\ref{prop:calibration-sample-size}. The diagnostic uses longer \(480\)-block streaming cases and treats the full \(120\)-block calibration split as the reference calibration set. For Criteo, the reference decision is the certified policy \texttt{pace\_i3\_q2}. With \(8\) calibration blocks, only \(25.0\%\) of resamples recover the same policy and decision state. At \(96\) calibration blocks, the recovery rate reaches \(1.0\), and the median shortlist size is \(3\). For KuaiRand, the reference decision is unresolved. The same decision state is recovered in \(97.9\%\) of resamples at \(8\) calibration blocks and \(100\%\) from \(12\) blocks onward. Policy-agreement is not reported for KuaiRand because the reference state is unresolved rather than a named policy.

The margin-condition curve is conservative. It records the condition from Proposition~\ref{prop:calibration-sample-size}. The maximum component quantile error must fall below half of the empirical decision margin. Criteo decision stability becomes perfect before this condition holds in every resample, which is expected. The proposition gives a reliable stability certificate, not an exact prediction of every stable empirical run.

\section{Proofs}

\subsection{Proof of Lemma~\ref{lem:quantile-concentration}}

Let \(p=1-\alpha\), and let \(F_n\) be the empirical distribution function of \(S_1,\ldots,S_n\). The split-conformal order statistic uses the rank
\[
r_n=\left\lceil (n+1)p\right\rceil.
\]
Equivalently, \(\widehat q=F_n^{-1}(\widehat p_n)\) with empirical quantile level \(\widehat p_n=r_n/n\). The conformal rank is slightly more conservative than the population level \(p\). Since
\[
(n+1)p\le r_n\le (n+1)p+1,
\]
we have
\[
p
\le
\widehat p_n
\le
p+\frac{2}{n}.
\]

By the density lower bound on \([q-\varepsilon,q+\varepsilon]\),
\[
F(q)-F(q-\varepsilon)
\ge
f_{\min}\varepsilon,
\qquad
F(q+\varepsilon)-F(q)
\ge
f_{\min}\varepsilon.
\]
Because \(q=F^{-1}(p)\), this implies
\[
F(q-\varepsilon)
\le
p-f_{\min}\varepsilon,
\qquad
F(q+\varepsilon)
\ge
p+f_{\min}\varepsilon.
\]

Consider the event
\[
\mathcal E_n
=
\left\{
\sup_x |F_n(x)-F(x)|
\le
\frac{f_{\min}\varepsilon}{4}
\right\}.
\]
On \(\mathcal E_n\),
\[
F_n(q-\varepsilon)
\le
F(q-\varepsilon)+\frac{f_{\min}\varepsilon}{4}
\le
p-\frac{3f_{\min}\varepsilon}{4}.
\]
Also,
\[
F_n(q+\varepsilon)
\ge
F(q+\varepsilon)-\frac{f_{\min}\varepsilon}{4}
\ge
p+\frac{3f_{\min}\varepsilon}{4}.
\]
If \(n\ge 8/(f_{\min}\varepsilon)\), then \(2/n\le f_{\min}\varepsilon/4\), so
\[
\widehat p_n
\le
p+\frac{2}{n}
\le
p+\frac{f_{\min}\varepsilon}{4}.
\]
Therefore,
\[
F_n(q-\varepsilon)
<
\widehat p_n
<
F_n(q+\varepsilon).
\]
By the definition of the empirical quantile, this implies
\[
q-\varepsilon
\le
\widehat q
\le
q+\varepsilon.
\]

It remains to control the probability of \(\mathcal E_n\). By the Dvoretzky--Kiefer--Wolfowitz inequality,
\[
\Pbb(\mathcal E_n^c)
\le
2\exp\left\{
-2n\left(\frac{f_{\min}\varepsilon}{4}\right)^2
\right\}
=
2\exp\left\{
-\frac{n f_{\min}^2\varepsilon^2}{8}
\right\}.
\]
Thus \(\Pbb(\mathcal E_n)\ge 1-\delta\) whenever
\[
n
\ge
\frac{8}{f_{\min}^2\varepsilon^2}
\log\frac{2}{\delta}.
\]
The additional rank-correction requirement \(n\ge 8/(f_{\min}\varepsilon)\) is implied by the same display after increasing the universal constant \(C\), using \(f_{\min}\varepsilon\le 1\) and \(\delta\in(0,1)\). Hence, for a universal constant \(C\),
\[
n
\ge
\frac{C}{f_{\min}^2\varepsilon^2}
\log\frac{2}{\delta}
\]
implies
\[
\Pbb\left(|\widehat q-q|\le \varepsilon\right)\ge 1-\delta.
\]

\subsection{Proof of Proposition~\ref{prop:calibration-sample-size}}

Apply Lemma~\ref{lem:quantile-concentration} to each calibrated decision component \(k\in\{V,g_1,\ldots,g_J\}\), with failure probability \(\delta/K_{\mathrm{comp}}\). If
\[
n_{\mathrm{cal}}
\ge
\frac{C}{f_{\min}^2\varepsilon^2}
\log\frac{2K_{\mathrm{comp}}}{\delta},
\]
then, by a union bound over the \(K_{\mathrm{comp}}\) components,
\[
\max_{k\in\{V,g_1,\ldots,g_J\}}
|\widehat q_k-q_k|
\le
\varepsilon
\]
with probability at least \(1-\delta\). This proves the first claim.

We now show how this event translates into the robust pacing quantities used by Eq.~\eqref{eq:robust-policy}. Define the population-calibrated robust value lower bound
\[
\mathcal V_0(\pi)
=
\widehat V(\pi)-q_V-\rho_V(\pi),
\]
and its empirical counterpart
\[
\widehat{\mathcal V}(\pi)
=
\widehat V(\pi)-\widehat q_V-\rho_V(\pi).
\]
On the quantile-concentration event,
\[
\left|
\widehat{\mathcal V}(\pi)-\mathcal V_0(\pi)
\right|
=
|\widehat q_V-q_V|
\le
\varepsilon
\qquad
\text{for all }\pi\in\PiC.
\]
Similarly, for constraint \(j\), define the population-calibrated robust constraint margin
\[
\mathcal M_{0,j}(\pi)
=
\widehat g_j(\pi)+q_{g_j}+\rho_{g,j}(\pi),
\]
and the empirical margin
\[
\widehat{\mathcal M}_{j}(\pi)
=
\widehat g_j(\pi)+\widehat q_{g_j}+\rho_{g,j}(\pi).
\]
Then, uniformly over \(\pi\in\PiC\),
\[
\left|
\widehat{\mathcal M}_{j}(\pi)-\mathcal M_{0,j}(\pi)
\right|
=
|\widehat q_{g_j}-q_{g_j}|
\le
\varepsilon.
\]
Thus every robust value lower bound and every robust constraint upper bound used by the empirical selector is within \(\varepsilon\) of the population-calibrated version, uniformly over the catalog.

It remains to verify the stated stability consequence. Let
\[
\Pi_0
=
\{\pi\in\PiC:\mathcal M_{0,j}(\pi)\le 0\text{ for all }j\}
\]
be the population-calibrated feasible catalog, and suppose \(\pi_0\in\Pi_0\) is the unique population-calibrated optimizer. Assume the population decision has margin larger than \(2\varepsilon\) in the following concrete sense:
\[
\mathcal M_{0,j}(\pi_0)\le -2\varepsilon
\quad\text{for all }j,
\]
every population-infeasible policy satisfies
\[
\max_j \mathcal M_{0,j}(\pi)>2\varepsilon,
\]
and every other population-feasible policy satisfies
\[
\mathcal V_0(\pi_0)-\mathcal V_0(\pi)>2\varepsilon.
\]
On the quantile-concentration event, \(\pi_0\) remains empirically feasible because
\[
\widehat{\mathcal M}_{j}(\pi_0)
\le
\mathcal M_{0,j}(\pi_0)+\varepsilon
\le
-\varepsilon
<
0.
\]
Any population-infeasible policy remains empirically infeasible because, for at least one \(j\),
\[
\widehat{\mathcal M}_{j}(\pi)
\ge
\mathcal M_{0,j}(\pi)-\varepsilon
>
\varepsilon
>
0.
\]
Therefore the empirical feasible set is contained in the population feasible set and contains \(\pi_0\). For any other empirically feasible policy \(\pi\), the value gap satisfies
\[
\widehat{\mathcal V}(\pi_0)-\widehat{\mathcal V}(\pi)
\ge
\mathcal V_0(\pi_0)-\mathcal V_0(\pi)-2\varepsilon
>
0.
\]
Hence \(\pi_0\) is also the unique empirical robust optimizer. If these \(2\varepsilon\) margins fail, then a perturbation of the calibrated quantiles within the guaranteed sampling error can change either feasibility or the robust value ordering. In that regime the calibration data support a set of near-optimal or near-feasible policies rather than a stable single-policy recommendation, which is the uncertainty-aware shortlist interpretation stated in the proposition.

\subsection{Proof of Theorem~\ref{thm:sharp-geometry}}

\noindent \textbf{Part I. Operational characterization.}
Fix the nominal world \((\widehat u,\widehat\lambda)\) and a forecast error \(e=z-\widehat z\). By Assumption~\ref{ass:affine}, for every policy \(\pi\in\PiC\), the inventory-dependent contribution to the evaluated pacing Lagrangian is affine in \(z\) with coefficient \(w_\pi(\widehat u,\widehat\lambda)\). Therefore,
\[
\mathcal L(\pi;(\widehat z+e,\widehat\tau,\widehat m),\widehat\lambda)
-
\mathcal L(\pi;(\widehat z,\widehat\tau,\widehat m),\widehat\lambda)
=
\left\langle
w_\pi(\widehat u,\widehat\lambda),e
\right\rangle .
\]
Taking absolute values and maximizing over the catalog gives
\[
\sup_{\pi\in\PiC}
\left|
\mathcal L(\pi;(\widehat z+e,\widehat\tau,\widehat m),\widehat\lambda)
-
\mathcal L(\pi;(\widehat z,\widehat\tau,\widehat m),\widehat\lambda)
\right|
=
\max_{\pi\in\PiC}
\left|
\left\langle
w_\pi(\widehat u,\widehat\lambda),e
\right\rangle
\right|
=
\Phi_{\PiC}(e;\widehat u,\widehat\lambda).
\]
This proves the first equality claim.

\noindent \textbf{Part II. Geometric representation.}
The support-function representation follows from the signed sensitivity set. By definition,
\[
\sigma_{\mathcal W_{\PiC}}(e)
=
\sup_{v\in\mathcal W_{\PiC}}
\langle v,e\rangle .
\]
Because a linear functional achieves its supremum over a finite convex hull at an extreme point,
\[
\sup_{v\in\mathcal W_{\PiC}}
\langle v,e\rangle
=
\max_{\pi\in\PiC}
\max\{\langle w_\pi,e\rangle,-\langle w_\pi,e\rangle\}
=
\max_{\pi\in\PiC}
|\langle w_\pi,e\rangle|
=
\Phi_{\PiC}(e;\widehat u,\widehat\lambda).
\]
Thus the decision-calibrated score is the support function of the convex hull of the signed policy sensitivities.

\noindent \textbf{Part III. Minimality characterization.}
This part formalizes a simple requirement. A valid scalar certificate must contain every signed sensitivity direction that can appear in the catalog. Let \(B\) be any lower-semicontinuous sublinear scalar certificate satisfying the uniform catalog-validity condition. Define its polar set
\[
\mathcal K_B
=
\{v:\langle v,e\rangle\le B(e)\ \text{for all }e\}.
\]
This set is closed and convex because it is an intersection of closed halfspaces. A standard polar representation for closed sublinear functions gives
\[
B(e)=\sigma_{\mathcal K_B}(e)
=
\sup_{v\in\mathcal K_B}\langle v,e\rangle .
\]
For completeness, the representation follows from Fenchel--Moreau duality. A closed sublinear function is equal to its biconjugate, and its convex conjugate is the indicator of \(\mathcal K_B\). Therefore \(B(e)=\sup_v\{\langle v,e\rangle-B^\star(v)\}=\sup_{v\in\mathcal K_B}\langle v,e\rangle\).

It remains to identify which sets \(\mathcal K_B\) are valid for the pacing catalog. The certificate condition and Assumption~\ref{ass:affine} imply that, for every policy \(\pi\) and every error \(e\),
\[
\left|
\left\langle
w_\pi(\widehat u,\widehat\lambda),e
\right\rangle
\right|
\le B(e).
\]
Equivalently,
\[
\langle w_\pi,e\rangle\le B(e)
\quad\text{and}\quad
\langle -w_\pi,e\rangle\le B(e)
\qquad
\text{for all }e.
\]
By the definition of \(\mathcal K_B\), this means \(w_\pi\in\mathcal K_B\) and \(-w_\pi\in\mathcal K_B\) for every \(\pi\). Since \(\mathcal K_B\) is convex,
\[
\mathcal W_{\PiC}(\widehat u,\widehat\lambda)\subseteq \mathcal K_B .
\]
Support functions are monotone under set inclusion, so
\[
\Phi_{\PiC}(e;\widehat u,\widehat\lambda)
=
\sigma_{\mathcal W_{\PiC}}(e)
\le
\sigma_{\mathcal K_B}(e)
=
B(e)
\qquad
\text{for all }e.
\]
This proves that every coherent uniformly valid scalar certificate must dominate the dual-weighted score.

Finally, suppose equality holds for all \(e\). Then \(\sigma_{\mathcal K_B}=\sigma_{\mathcal W_{\PiC}}\). Closed convex sets are uniquely determined by their support functions, so \(\mathcal K_B=\mathcal W_{\PiC}\). If \(\mathcal K_B=\mathcal W_{\PiC}\), then \(B=\sigma_{\mathcal K_B}=\sigma_{\mathcal W_{\PiC}}=\Phi_{\PiC}\). This proves that the dual-weighted score is the unique minimal coherent scalar certificate for the catalog.

\noindent \textbf{Part IV. Conformal validity.}
For the calibration claim, condition on the training data, the fitted forecasting model, the nominal pacing solutions, and the dual prices used to compute the scores. Under Assumption~\ref{ass:exchange}, the calibration scores
\[
S_1^{\mathrm{dc}},\ldots,S_n^{\mathrm{dc}}
\]
and the test-block score
\[
S_{n+1}^{\mathrm{dc}}
=
\max_{\pi\in\PiC}
\left|
\left\langle
w_\pi(\widehat u,\widehat\lambda),
z-\widehat z
\right\rangle
\right|
\]
are exchangeable. Split conformal prediction therefore gives the rank guarantee
\[
\Pbb\!\left(
S_{n+1}^{\mathrm{dc}}
\le
\widehat q^{\mathrm{dc}}_{1-\alpha}
\right)
\ge
1-\alpha,
\]
where \(\widehat q^{\mathrm{dc}}_{1-\alpha}\) is the order statistic in Eq.~\eqref{eq:dc-quantile}. By Assumption~\ref{ass:affine}, the inventory-dependent Lagrangian error for policy \(\pi\) is exactly
\[
\mathcal L(\pi;(z,\widehat\tau,\widehat m),\widehat\lambda)
-
\mathcal L(\pi;(\widehat z,\widehat\tau,\widehat m),\widehat\lambda)
=
\left\langle
w_\pi(\widehat u,\widehat\lambda),
z-\widehat z
\right\rangle.
\]
Taking the maximum over \(\pi\in\PiC\) gives the event in the theorem. If the downstream functions are differentiable rather than affine, the same proof applies to the first-order term and the bound gains the usual Taylor remainder
\[
\frac{L_2}{2}\|z-\widehat z\|^2
\]
whenever the Hessian norm is bounded by \(L_2\) on the relevant uncertainty neighborhood.

\subsection{Proof of Theorem~\ref{thm:separation}}

Let \(P_{\mathcal S}\) and \(P_{\mathcal N}\) be the orthogonal projections onto \(\mathcal S\) and \(\mathcal N\). By assumption, every catalog sensitivity vector lies in \(\mathcal S\). Therefore, for any forecast error \(e=e_{\mathcal S}+e_{\mathcal N}\) and any policy \(\pi\in\PiC\),
\[
\langle w_\pi,e\rangle
=
\langle w_\pi,e_{\mathcal S}\rangle
+
\langle w_\pi,e_{\mathcal N}\rangle
=
\langle w_\pi,e_{\mathcal S}\rangle,
\]
because \(w_\pi\perp e_{\mathcal N}\). Maximizing over \(\pi\in\PiC\) gives
\[
\Phi_{\PiC}(e;\widehat u,\widehat\lambda)
=
\max_{\pi\in\PiC}|\langle w_\pi,e\rangle|
=
\max_{\pi\in\PiC}|\langle w_\pi,e_{\mathcal S}\rangle|
=
\Phi_{\PiC}(e_{\mathcal S};\widehat u,\widehat\lambda).
\]
This proves that the score ignores nuisance directions orthogonal to all pacing sensitivities. In contrast, by the Pythagorean theorem,
\[
\|e\|_2^2
=
\|e_{\mathcal S}\|_2^2+\|e_{\mathcal N}\|_2^2,
\]
so a generic Euclidean residual score pays for both decision-relevant and nuisance forecast error.

We now give the explicit separation construction. Fix an integer \(m\ge 1\) and a scale \(\sigma>0\). Let the inventory forecast live in \(\R^{m+1}\). Let the pacing catalog contain a single policy with sensitivity vector
\[
w=e_1=(1,0,\ldots,0).
\]
The decision-relevant subspace is \(\mathcal S=\mathrm{span}\{e_1\}\), and the nuisance subspace is the span of the remaining \(m\) coordinate vectors. Let each calibration block and the test block have forecast error
\[
e=(\xi_0,\sigma \xi_1,\ldots,\sigma \xi_m),
\]
where \(\xi_0,\xi_1,\ldots,\xi_m\) are independent Rademacher random variables. The calibration and test errors are exchangeable because they are independent draws from the same distribution.

The decision-calibrated score is
\[
\Phi_{\PiC}(e)
=
|\langle e_1,e\rangle|
=
|\xi_0|
=
1.
\]
Thus every calibration score equals \(1\), and the split-conformal radius based on the decision-calibrated score is
\[
q^{\mathrm{dc}}_{1-\alpha}=1.
\]
This radius has coverage one and, by Theorem~\ref{thm:sharp-geometry}, is the sharp radius for the catalog Lagrangian forecast error.

For a generic Euclidean residual score,
\[
S^{\mathrm{gen}}(e)=\|e\|_2
=
\sqrt{1+m\sigma^2}.
\]
Every generic calibration score is also equal to this value, because the Rademacher signs change direction but not Euclidean length. Therefore,
\[
q^{\mathrm{gen}}_{1-\alpha}
=
\sqrt{1+m\sigma^2}.
\]
Therefore,
\[
\frac{q^{\mathrm{gen}}_{1-\alpha}}{q^{\mathrm{dc}}_{1-\alpha}}
=
\sqrt{1+m\sigma^2}.
\]
For any target gap \(A>1\), choose \(m\) and \(\sigma\) such that \(m\sigma^2\ge A^2-1\). Then \(q^{\mathrm{gen}}_{1-\alpha}\ge A q^{\mathrm{dc}}_{1-\alpha}\). The generic certificate is valid, but its excess slack is caused entirely by nuisance inventory directions that do not change any catalog policy's downstream Lagrangian value. This proves the claimed high-dimensional separation.

\subsection{Proof of Theorem~\ref{thm:robust-certificate}}

Let \(E\) be the simultaneous event in Assumption~\ref{ass:responsecoverage}. Then
\[
\Pbb(E)
\ge
1-\alpha-\alpha_\tau-\alpha_m.
\]
Work on this event. For every policy \(\pi\) and constraint \(j\),
\[
g_j(\pi;u^\star)
\le
\widehat g_j(\pi)+\widehat q^{g_j}_{1-\alpha_{\mathrm{comp}}}+\rho_{g,j}(\pi).
\]
The selected policy \(\widehat\pi\) satisfies the robust constraints in Eq.~\eqref{eq:robust-policy}, so
\[
\widehat g_j(\widehat\pi)+\widehat q^{g_j}_{1-\alpha_{\mathrm{comp}}}+\rho_{g,j}(\widehat\pi)
\le 0.
\]
Combining the two inequalities yields \(g_j(\widehat\pi;u^\star)\le 0\) for all \(j\).

For regret, define
\[
\Delta_V(\pi)
=
\widehat q^V_{1-\alpha_{\mathrm{comp}}}+\rho_V(\pi).
\]
On the same event,
\[
V(\pi;u^\star)\le \widehat V(\pi)+\Delta_V(\pi),
\qquad
V(\pi;u^\star)\ge \widehat V(\pi)-\Delta_V(\pi).
\]
Because Eq.~\eqref{eq:robust-policy} is solved to error \(\varepsilon_{\mathrm{opt}}\),
\[
\widehat V(\widehat\pi)-\Delta_V(\widehat\pi)
\ge
\widehat V(\pi^\star_{\mathrm{rob}})-\Delta_V(\pi^\star_{\mathrm{rob}})
-\varepsilon_{\mathrm{opt}}.
\]
Therefore,
\[
\begin{aligned}
V(\pi^\star_{\mathrm{rob}};u^\star)-V(\widehat\pi;u^\star)
&\le
\widehat V(\pi^\star_{\mathrm{rob}})+\Delta_V(\pi^\star_{\mathrm{rob}})
-
\left\{\widehat V(\widehat\pi)-\Delta_V(\widehat\pi)\right\} \\
&\le
2\Delta_V(\pi^\star_{\mathrm{rob}})+\varepsilon_{\mathrm{opt}} \\
&\le
2\sup_{\pi\in\PiC}\Delta_V(\pi)+\varepsilon_{\mathrm{opt}}.
\end{aligned}
\]
This proves the claimed bound.

\subsection{Proof of Theorem~\ref{thm:slack}}

Because the constraint contribution is affine in the inventory forecast error, the realized constraint value under perturbation \(e\) is
\[
g_j(\pi;e)
=
\widehat g_j(\pi)+\langle a_{\pi,j},e\rangle.
\]
The uncertainty set contains a perturbation whose inner product with \(a_{\pi,j}\) has magnitude \(\delta\). Choose the sign that increases the constraint value, so that
\[
\langle a_{\pi,j},e\rangle=\delta.
\]
If the point-forecast margin is smaller than \(\delta\), then
\[
\widehat g_j(\pi)+\delta>0.
\]
Thus \(g_j(\pi;e)>0\) for a plausible future in the same decision-calibrated uncertainty set. This proves that slack at least \(\delta\) is necessary for certification over that set. The statement is sharp in the affine case because slack equal to the supremum of \(|\langle a_{\pi,j},e\rangle|\) is also sufficient for that constraint.

\subsection{Proof of Proposition~\ref{prop:catalog}}

Let \(\pi^\star\in\arg\inf_{\pi\in\Pi}R(\pi)\).
Because \(\PiC\) is an \(\eta\)-net under \(d_{\mathrm{pace}}\), there exists \(\pi_C\in\PiC\) with
\[
d_{\mathrm{pace}}(\pi_C,\pi^\star)\le \eta.
\]
Lipschitzness gives
\[
R(\pi_C)\le R(\pi^\star)+L_\Pi \eta.
\]
Since the best catalog policy has risk no larger than \(R(\pi_C)\),
\[
\inf_{\pi\in\PiC}R(\pi)
\le
R(\pi_C)
\le
\inf_{\pi\in\Pi}R(\pi)+L_\Pi\eta.
\]

\end{document}